\documentclass[conference]{IEEEtran}
\IEEEoverridecommandlockouts
\usepackage{array}
\usepackage{times}
\usepackage{soul}
\usepackage{url}
\usepackage[bookmarks=false, hidelinks]{hyperref}
\usepackage[utf8]{inputenc}
\usepackage[small]{caption}
\usepackage{graphicx}
\usepackage{subcaption}
\usepackage{amsmath}
\usepackage{amsfonts}
\usepackage{booktabs}
\usepackage{algorithm}
\usepackage{algorithmic}
\usepackage{dblfloatfix}
\usepackage{multirow}
\usepackage{caption} 
\newcommand{\bftab}{\fontseries{b}\selectfont}

\usepackage[utf8]{inputenc}
\usepackage{amsthm,amsmath}
\usepackage{xcolor}

\newcommand{\etal}{\textit{et al}.}
\newcommand{\ie}{\textit{i}.\textit{e}.}
\newcommand{\eg}{\textit{e}.\textit{g}.}

\title{Learning Classifiers on Positive and Unlabeled Data with Policy Gradient\\
}

\author{    
\IEEEauthorblockN{
    	Tianyu Li\IEEEauthorrefmark{1}\thanks{Correspondence to: tianyu.li@rakuten.com}, Chien-Chih Wang\IEEEauthorrefmark{1}, Yukun Ma\IEEEauthorrefmark{2}, Patricia Ortal\IEEEauthorrefmark{1}, Qifang Zhao\IEEEauthorrefmark{1}, Bj\"orn Stenger\IEEEauthorrefmark{1}, Yu Hirate\IEEEauthorrefmark{1}}
    \IEEEauthorblockA{\IEEEauthorrefmark{1}Rakuten Institute of Technology, Tokyo
    }
    \IEEEauthorblockA{\IEEEauthorrefmark{2}AIR Labs, Continental Automotive Group, Singapore
    }
}

\begin{document}
\maketitle

\begin{abstract}
Existing algorithms aiming to learn a binary classifier from positive (P) and unlabeled (U) data
 require estimating the class prior or label noise ahead of building a classification model. However, the estimation and classifier learning are normally conducted in a pipeline instead of being jointly optimized.
In this paper, we propose to alternatively train the two steps using reinforcement learning. Our proposal adopts a policy network to adaptively make assumptions on the labels of unlabeled data, while a classifier is built upon the output of the policy network and provides rewards to learn a better policy.
The dynamic and interactive training between the policy maker and the classifier can exploit the unlabeled data in a more effective manner and yield a significant improvement in terms of classification performance.
Furthermore, we present two different approaches to represent the actions taken by the policy.
The first approach considers continuous actions as soft labels, while the other uses discrete actions as hard assignment of labels for unlabeled examples.
We validate the effectiveness of the proposed method on two public benchmark datasets as well as one e-commerce dataset. The results show that the proposed method is able to consistently outperform state-of-the-art methods in various settings.

\end{abstract}


\section{Introduction}
PU learning refers to the problem of learning from a dataset where only a subset of examples are positively labeled and the rest are not annotated at all.
It is a critical task due to its prevalence in various real-world applications~\cite{Ward2009:PresenceonlyDA,li2010:remote-sensing,mordelet2010:svm}. In many common situations only positive data are available, for instance, an e-commerce website may only record users who have clicked on advertisements or purchased items.
Meanwhile, it is not possible to simply assume that unlabeled instances are negative. Another example is diagnosis systems that predict whether or not a patient has a certain disease. To build such systems, the already diagnosed patients are naturally treated as positives. Yet, we cannot infer that all undiagnosed patients are not suffering from the disease.

The process of PU learning is conventionally done in two steps: (1) identify likely negative samples from unlabeled data and (2) perform traditional supervised learning on labeled positives and reliable negatives (N) \cite{Liubing2002:icml, liubing2003:icdm,xiaoli2003:ijcai,Nguyen:ijcai2011}. 
More recent research focuses on estimating label noise in the unlabeled dataset or the class prior of the training dataset, and then exploit the estimated values during the classifier training.
The work in~\cite{elkan-noto:pulearning2008} made a notable breakthrough by modeling each unlabeled data point as a mix of both positive and negative classes. 
In the case that the class prior is known, the learning on P and U can be reformulated as a cost-sensitive classification problem~\cite{elkan2001:cslearning}.
The work in~\cite{plessis:pu2014} introduces a risk estimator that exploits non-convex loss functions, \eg, the ramp loss, to cancel estimation bias. A more general estimator which is unbiased and convex by utilizing different loss functions for positive and unlabeled examples is further proposed in~\cite{plessis:pu2015}. 
Incorrectly labeled examples can be removed from the original PU dataset based on noisy label prediction, allowing for training a better classification model~\cite{northcutt2017:rankpruning}.

The class prior and label noise rate are essential to existing PU learning approaches and have to be estimated before training the classifier. 
However, the prior distribution of labels or the possible mislabeled examples in the unlabeled dataset are unknown in typical real-world scenarios~\cite{Blanchard2010:novelty,Natarajan2013:LNL,Jain2016:NonparametricSL}. Consequently, the resulting classifier is affected by the estimation accuracy of prior and label noise. 
Moreover, the two-step process is unidirectional, \ie, there is no feedback from the classification to the prior and label noise estimation.
As a result, the pipeline of existing methods leads to non-optimal classification on PU datasets.

This paper proposes a reinforcement learning framework to jointly estimate the labels of unlabeled data and learn a binary classifier.
The whole framework can be trained in an end-to-end fashion.
Our framework, named \emph{policyPU}, consists of two components: a policy network and a classifier. The policy network learns to infer label assignment for the unlabeled data,
while the classifier is trained using the data and label estimates.  
The policy network, serving as an agent, formulates the input attribute vector as state and receives rewards from the classifier to update with the policy gradient. It gradually improves its decision making and generates a more accurate output that maximizes the expected reward from the classifier.
We present two variants of our framework in terms of learning different policies for unlabeled data. 
The classifiers use distinct objective functions accordingly. 
In the first approach, we assume that U data is a combination of P and N~\cite{elkan-noto:pulearning2008}. The policy network produces continuous action values within $(0, 1)$ as soft labels.
The second approach applies discrete actions of the policy network as hard label assignments, which allows us to use standard supervised learning on the complete dataset. The hard assignment can be obtained by simply thresholding continuous label values.


Regardless of the different strategies, the policy network and the classifier are trained iteratively to learn a policy which makes correct assumptions for U data, and eventually a classifier that fully exploits both P and U so that it has a better generalization ability.



The technical contributions of this paper are summarized as follows:
\begin{enumerate}
    \item We propose a policy network for explicitly inferring the label assignment of unlabeleled data through a dynamic interaction with the classifier. Compared to existing methods that estimate unlabeled examples beforehand, we exploit the underlying structure of unlabeled data more effectively by taking  the targeted classifier performance into consideration.
    \item 
    Two approaches are presented for applying the outcome of policy network differently. The  classifiers are trained with either continuous or discrete actions accordingly. 
    Especially, the continuous actions allow a classifier to explore unlabeled instance as a mixture of positive and negative.
    \item We conduct comprehensive experiments and show that the classifiers learned by our framework yield consistent improvements in terms of accuracy, the area under the ROC curve (ROC$\_$AUC), and the area under the precision-recall curve (PR$\_$AUC) on three datasets.
\end{enumerate}

\section{PU Learning Settings}

PU learning is to build a classifier from positive and unlabeled training data. Although the inputs to PN and PU learning are different, they share the same goal, namely to apply the resulting classifier to distinguish positive and negative samples in test data. 

Let $x$ be the feature vector of a sample, $y \in \{0,1\}$  its true class label and $s \in \{0, 1\}$ its status of being labeled or not. We represent a PU dataset as a set of triplets $\langle x, y, s \rangle$, which consists of a set of labeled examples $ \langle x, s=1 \rangle$ and a set of unlabeled examples $ \langle x, s=0 \rangle$. Since only positive examples are labeled, $s=1$ indicates $y=1$. For $s=0$, either $y=1$ or $y=0$ could be true. 
A general assumption for current PU learning methods is the Selected Completely At Random (SCAR) assumption. It assumes that all labeled samples are selected completely at random from the entire positive example set, indicating that the $s$ label and the attribute $x$ are conditionally independent from the true class $y$ \cite{elkan-noto:pulearning2008}. It is formally stated as:
\begin{equation}
    p(s=1|x, y=1) = p(s=1|y=1).
\end{equation}

The value of $c = p(s=1|y=1)$ is the constant probability of a positive example being labeled, referred as label frequency \cite{Bekker2018EstimatingTC}. Elkan \cite{elkan-noto:pulearning2008} proves the following property between class prior and label frequency $c$:
\begin{equation}
\label{eq:alpha_c}
    p(y=1|x) = p(s=1|x) / c.
\end{equation}
Equation (\ref{eq:alpha_c}) has been significant for existing PU learning algorithms. 

\section{Learning Classifiers on PU datasets via Policy Gradient}
\subsection{Overview}
\begin{figure}
  \centering
  \includegraphics[width=\columnwidth]{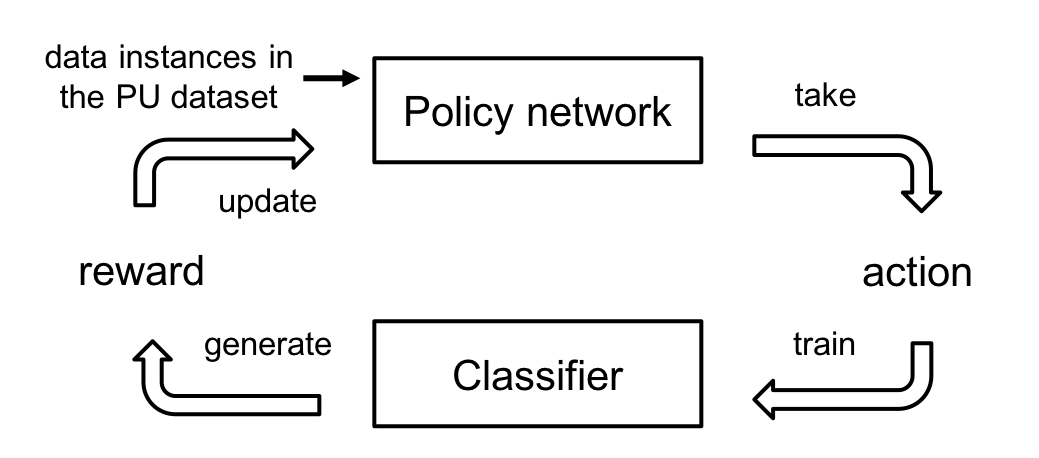}
  \caption{{\bf The diagram of the proposed reinforcement learning framework.} \it 
  The policy network takes actions on the input feature vectors.
  The training data and their actions from the policy are applied to learn a classification model.
  The policy receives the predicted class label probabilities by the classifier as rewards to update parameters with policy gradient.}
  \label{fig:framework_diagram}
\end{figure}
This paper presents a framework, \emph{policyPU}, in which a classifier is learned from positive and unlabeled examples via interacting with a policy network, as shown in Fig.~\ref{fig:framework_diagram}. 
Given a PU dataset, we explore how to learn a more accurate classifier by exploiting the unlabeled examples. 
Inspired by reinforcement learning \cite{Lillicrap2016:ContinuousCW}, our \emph{policyPU} dynamically adjusts its assumptions to U data after making decisions and receiving rewards from the classifier.
Thus, it is able to learn a classifier given a PU dataset in an end-to-end fashion.

To be more specific, the policy network acts as an \emph{agent}, while the target classifier and the PU dataset serve as the \emph{environment} in our reinforcement learning setting.
The attribute vector of data instances in the training dataset is \emph{state}, and the \emph{action} represents how the data example is used for classifier training. In practice, a sequence of mini-batches in our training process is formulated as \emph{trajectory}. Hence, the interaction between the agent and environment is as follows: the agent (the policy network) takes actions (label assignment) with input states (attribute vectors), and the classifier determines rewards for the agent to update its policy.
We denote two different approaches to learn policies as \emph{Weighter} and \emph{Separator}, respectively. 
In the rest of this Section, we first describe the policy networks, and then elaborate on the reward design. This is followed by the description of the classifiers. Finally, the iterative training procedure is presented.

\subsection{Policy Networks for PU datasets}
In order to learn a generalized classification model with P and U data, we would like to make better use of unlabeled examples. We formulate this solution-seeking process as a reinforcement learning task
 by defining the feature vector $x$ as \emph{state} and the output of the policy network as  \emph{action} for input $x$. The goal is then to learn a policy, $\pi_\Theta = p(a|x)$, that infers how each data sample in the training dataset contributes to the classifier. Let $\mathbb{P}$ be the labeled example set, and $\mathbb{U}$ the unlabeled example set. The objective of the policy network is to generate actions for data instances that maximize its expected reward:
\begin{equation}
    J(\Theta) = \sum_{x \in \mathbb{P} \cup \mathbb{U}} \pi_\Theta(a|x) \mathit{R}(x, a),
\label{eq: policy_obejctive}
\end{equation}
where $\mathit{R}(x, a)$ is the reward by the data instance with feature vector $x$ after taking action $a$. The reward $\mathit{R}(x, a)$ is defined as the class label probability given by, $\mathbf{F}_\Phi$, the classifier in our framework.

\subsection{Classification Coherence Rewards}
The core of our proposed framework is the learning of an effective policy to infer the labels of unlabeled data. To achieve this goal, we seek the feedback from the on-going classifier training process. 
The intuition behind our reward design is that eventually a good policy will be coherent with the classifier, and this coherence is valid for all data instances and across mini-batches in our framework setting.
More specifically, we leverage the probability of positive examples predicted by the classifier as references to decide whether an unlabeled data instance may be a plausible P or N, and define the reward function as:
\begin{equation}
\label{eq:return_func}
    R(x, a)= 
    \begin{cases}
        \hat{y}, & \text{if } x \in \mathbb{P}, \\
        \hat{y}, & \text{if } x \in \mathbb{U} \text{ and }  \hat{y} \geq \emph{threshold}, \\
        1-\hat{y}, & \text{if } x \in \mathbb{U} \text{ and }  \hat{y} < \emph{threshold},
    \end{cases}
\end{equation}
where $\hat{y}=\mathbf{F}_\Phi(x)$ is the predicted probability by the classifier for input vector $x$, and the \emph{threshold} is used as a reference for unlabeled examples.
We use the minimum class label prediction of positive examples as our first threshold value denoted as, $ thresh_{min} = \min_{x\in\mathbb{P}}\mathbf{F}_\Phi(\hat{y}|x)$.
Those unlabeled examples with $\hat{y}$  larger than this value, together with all positive examples are used to computer the threshold value in Equation (\ref{eq:return_func}):
\begin{equation}
\label{eq:threshold_calculation}
    \emph{threshold} = \mathbb{E}\left[\sum_{x\in\mathbb{P \cup U'}}\mathbf{F}_\Phi(\hat{y}|x)\right],
\end{equation}
where $\mathbb{U'} = \{\,x \mid \mathbf{F}_\Phi(\hat{y}|x) \geq thresh_{min}\,\}$.

For  unlabeled examples of which $\hat{y} \geq \emph{threshold}$, we trust the classifier's predicted label $y=1$.
Hence, we use their $\hat{y}$ as reward, the same as that of positive examples in the training dataset, otherwise $1-\hat{y}$ as they are predicted to be negative.

The goal of the policy learning is to optimize parameter $\Theta$ to output actions that can maximize the expected reward on both the labeled and unlabeled data \cite{Feng2018:ReinforcementLF}.
We update the policy by mini-batch training in practice.
Since the policy maker in our framework gets instantaneous rewards from the classifier after each mini-batch, we apply the REINFORCE algorithm to maximize $J(\Theta)$ \cite{williams1992:reinforce, liyan2018:adversarial, Zhang2018:LearningSR}. Its gradient is computed based on policy gradient theorem \cite{sutton1999:reinforce} as follows:
\begin{equation}
    \nabla_\Theta J(\Theta) = \mathbb{E}_\Theta \left[\nabla_{\pi_\Theta}
    log(\pi_\Theta(a|x))\mathit{R}(x, a)\right],
\end{equation}
where $\mathit{R}(x, a)$ is obtained by Equation (\ref{eq:return_func}) for both labeled data $\langle x, s=1 \rangle$ and unlabeled data $\langle x, s=0 \rangle$.

Let the batch size be $m$, then the parameter $\Theta$ of the policy network is updated via:
\begin{equation}
\label{eq:policy_minibatch_update}
    \Theta \leftarrow \Theta + \eta \frac{1}{m} \sum_{i=1}^{m} \nabla_{\pi_\Theta}
    log(\pi_\Theta(a_i|x_i))\mathit{R}(x_i, a_i),
\end{equation}
where $\eta$ is the learning rate.

\subsection{Classifiers on PU datasets}

With the actions from the policy network, all training data are used to learn a classifier. The classifier is denoted as $\mathbf{F}_\Phi$, where $\mathbf{F}$ is a differentiable classification model parameterized by $\Phi$.

In \emph{Weighter}, the policy network outputs continuous actions, $a \in (0,1)$, as soft labels for unlabeled data instances.
They are used to compute the weighted cost function of the corresponding classifier.
The network learns the weighting policy that  maximizes the classifier reward.
It generates weighted data and receives rewards to update its parameter $\Theta$ to improve its judgement. The classifier in \emph{Weighter} follows the idea from prior work that each unlabeled sample is a combination of a positive example with weight $w_i$ and a negative example with weight $1-w_i$, where $w_i \in (0,1)$ is a continuous real value and $x_i$ is the feature vector of example $i$ \cite{elkan-noto:pulearning2008}.
The objective function to learn the classifier in \emph{Weighter} is:

\begin{flalign}
\label{eq:puclassifier_objective_func}
\begin{split}
    \mathbf{L}(\Phi)= - \mathbb{E}_\Phi & [\sum_{x \in \mathbb{P}}log(\mathbf{F}_\Phi(\hat{y}|x))]\\
    - \mathbb{E}_\Phi & [\sum_{x \in \mathbb{U}}( w \, log(\mathbf{F}_\Phi(\hat{y}|x))  \\
    & \hphantom{-\sum_{x \in \mathbb{U}}} + (1-w) \, log(1-\mathbf{F}_\Phi(\hat{y}|x))) ],
\end{split}
\end{flalign}
where $w$ is the weight for an unlabeled example to be positive and $\hat{y}$ is the predicted probability given feature vector $x$. Here, the action value $a=\pi_\Theta(x)$ for $x$ is directly used as $w$ in the cost function. 
The loss function is minimized to learn the model $\mathbf{F}_\Phi$, which not only discriminates labeled and unlabeled samples, but also correctly identifies the contribution of unlabeled data to the model training.

The action $a \in \{0, 1\}$ of \emph{Separator} is a hard assignment, indicating whether an unlabeled data instance is assigned as positive or negative.  
The hard assignment can be seen as the soft labels thresholded with a value, set to $0.5$ in our experiments. A data sample with action value larger than this threshold is assigned as positive, otherwise negative. The learned policy here aims to identify those data examples in U that can be directly put into the labeled example set. The corresponding classifier is trained on generated P and N data via minimizing a cross-entropy loss function.

As described, according to the discrete actions of the policy network, some unlabeled data are selected as P, denoted as $\mathbb{P'}$, and the rest are used as negative, denoted as $\mathbb{N'}$. 
The classifier for \emph{Separator} is a standard supervised model with the cross-entropy cost function:
\begingroup\makeatletter\def\f@size{9}\check@mathfonts
\begin{equation}
\label{eq:separator_objective_func}
\mathbf{L}(\Phi)=-\mathbb{E}_\Phi[\sum_{x\in\mathbb{P \cup P'}}log(\mathbf{F}_\Phi(\hat{y}|x))+{\sum_{x\in \mathbb{N'}}}log(1-\mathbf{F}_\Phi(\hat{y}|x))].
\end{equation}
\endgroup

\subsection{The iterative training between the policy and the classifier}

The policy network and the classifier  interact in the following way: unlabeled examples in the training dataset are input to the policy network, which outputs either discrete actions or continuous-valued actions. The classifier takes the positive examples and unlabeled examples with their corresponding action values as input, generates class label probability for each data sample. The prediction results of the classifier are used as rewards for the policy maker. During the training, whether an unlabeled data example should be put in the positive dataset or how it is shared as both positive and negative simultaneously is learned and adjusted dynamically.

Our framework trains deep neural networks, Convolutional Neural Networks (CNNs) and Multilayer Perceptrons (MLPs), as classification models and policy networks.
The neural network training is done using mini-batch training. In each epoch, we randomly shuffle the data to create a trajectory of mini-batches.
Given a dataset that consists of both labeled and unlabeled data, a mini-batch with $m$ instances is first randomly sampled from the training dataset to obtain feature vectors 
$\{x_0, x_1, ..., x_{m-1}\}$ as states. For each state, an action $a$ is then taken by the policy network $\pi_\Theta$. 
The generated $\{x_0, a_0, x_1, a_1 ..., x_{m-1}, a_{m-1}\}$ is fed to train the classifier $\mathbf{F}_\Phi$. The class label prediction results for data instances in the current mini-batch are in return applied as reward to update the policy. They are jointly optimized and their parameters, $\Phi$ and $\Theta$, are updated every mini-batch. 

In \emph{Weighter}, the weight of a labeled example is set as $1$ for the objective function in Equation (\ref{eq:puclassifier_objective_func}), and that of an unlabeled example is the sampled continuous-valued action. In \emph{Separator}, those labeled instances are used as P in classification directly, while the unlabeled instances are separated based on their sampled actions.
\begin{algorithm}[tb]
\caption{\emph{policyPU} learning}
\label{alg:algorithm}
\textbf{Input}: a training dataset consists of P and U data \\
\textbf{Parameter}: $\Theta$; $\Phi$; batch size $m$; iteration number $n\_epochs$; policy update frequency $k$ \\
\textbf{Output}: $\pi_\Theta$; $\mathbf{F}_\Phi$
\begin{algorithmic}
\STATE Initialize target policy: $\Theta' \leftarrow \Theta$
\FOR {$epoch = 1$ to $n\_epochs$}
    \STATE shuffle to create mini-bathes
    \FOR{each mini-batch}
        \STATE Sample action $a_i \sim \pi_\Theta'$ from target policy for $x_i$;
        \STATE Minimize Equation (\ref{eq:puclassifier_objective_func})/(\ref{eq:separator_objective_func}) to learn the classifiers 
        \STATE using the generated $\{x_0, a_0, x_1, a_1 ..., x_{m-1}, a_{m-1}\}$;
        \STATE Predict class probability $\hat{y}_i = \mathbf{F}_\Phi(x_i)$ for $x_i$;
        \STATE Calculate the \emph{threshold} via Equation (\ref{eq:threshold_calculation});
        \STATE Get $R(x_i, a_i)$ via Equation (\ref{eq:return_func}) for taking action $a_i$;
        \STATE Update policy parameter $\Theta$ using Equation (\ref{eq:policy_minibatch_update})
    \ENDFOR
    \IF{$k \mid epoch$}
    \STATE Update target policy: $\Theta' \leftarrow \Theta$
    \ENDIF
\ENDFOR
\end{algorithmic}
\end{algorithm}
The classifier in the framework is built upon the P data and the processed U data. It updates parameter $\Phi$ and predicts class label probabilities, $p=\mathbf{F}_\Phi(\hat{y}|x)$, for all training data instances. They are the rewards for the policy network to update $\Theta$.

To reduce high variance of the returned reward, we adopt a target policy network for sampling actions. The network is updated every $k$ epochs in our experiment. The whole training process is illustrated in Algorithm \ref{alg:algorithm}. Besides, a pre-training step, that simply use unlabeled examples as negative, is applied before the interactive learning to stabilize the process as well. First, the classifier is trained using P and U data directly via a few epochs, then the policy network is also pre-trained with several iterations using the prediction outcome of the classifier.

\section{Related Work}

Elkan \etal \cite{elkan-noto:pulearning2008} first 
shows that the output probability of a classifier which predicts $p=(s=1|x)$ can be adjusted via Equation (\ref{eq:alpha_c}) to get $p=(y=1|x)$ under the SCAR assumption.
It also proposes that each unlabeled example can be utilized as a combination of a positive with weight $w(x)$ and a negative with complementary $1-w(x)$, where $w(x) = \frac{1-c}{c} \frac{p(s=1|x)}{1-p(s=1|x)}$, and c is the label frequency. Besides, similarity based method is proposed to associate ambiguous data instances with two similarity weights indicating their resemblance to positive and negative examples, respectively \cite{Xiao2011:similarity}.

The research in \cite{plessis:pu2014} proposes an unbiased risk estimator to learn classifiers. 
Let $g$ be a decision function, $l$ be a loss function and $\alpha$ be the class prior. The risk of $g$ in the learning on positive and negative examples is:
\begin{equation}
    \label{eq:pn_risk}
    \mathcal{R}(g)=\alpha \mathbb{E}_{p} [l(g(x))] + (1-\alpha) \mathbb{E}_{n}[l(-g(x))].
\end{equation}
Given the key observation that $(1-\alpha)\mathbb{E}_n[l(-g(x))]$ can be approximated using $\mathbb{E}_u[l(-g(x))]- \alpha\mathbb{E}_p[l(-g(x))]$, the Equation (\ref{eq:pn_risk}) is rewritten as:
\begin{equation}
    \mathcal{R}(g)=\alpha \mathbb{E}_p[l(g(x))] - \alpha \mathbb{E}_p[l(-g(x))] + \mathbb{E}_u[l(-g(x))],
\end{equation}
for PU learning.
But this method requires a loss function to satisfy $\mathit{l}(x)+\mathit{l}(-x) =1$, \eg, the ramp loss.
To apply this unbiased risk estimator to train deep neural networks, a non-negative variation is proposed to alleviate overfitting and to be implemented for large scale training data by stochastic optimization \cite{Kiryo2017PositiveUnlabeledLW}. Another recent research \cite{Shi2018:PUL} also converts PU learning to risk minimization problem, in which it adopts different loss functions for P and U, respectively \cite{gao2016:riskmini}. The proposed method further shows that its risk minimization in the presence of noisy negative data can be turned into the estimation of the centroid of negative examples.


As knowing the label noise rate or the class prior of training dataset simplifies PU learning greatly, plenty of methods have been proposed to directly conduct estimation from PU datasets \cite{plessis2014:pe, Bekker2019:survey}. 
Several mixture proportion estimation methods are proposed recently, in which the unlabeled data are considered as a mixture of positive distribution and an unknown negative distribution \cite{scott15:pmlr}.
\cite{Jain2016:EstimatingTC} proposes to use kernels to model distributions, and the class prior is the sum of weights that represent how much the positive distribution contributes to each kernel.
\cite{Ramaswamy2016:MPE} further proposes a similar algorithm but using the distance between kernel embeddings to find the optimal weights.
More recently, \cite{Bekker2018EstimatingTC} proposes a decision tree based approach to estimate label frequency, then use Equation (\ref{eq:alpha_c}) as a medium to get class prior. 
RankPruning \cite{northcutt2017:rankpruning} tempts to guess incorrectly labeled examples in the training datasets, then prune those examples with low predicted probabilities after ranking them. Then, a classifier with weighted cost function is learned on the pruned training datasets. 


\section{Experiments}

\begingroup
\setlength{\tabcolsep}{1.5pt}
\begin{table}[t!]
\caption{{\bf Benchmark datasets.} \it The number of data instances in train and test, the feature vector dimension, also the architecture of the classifiers and the policy networks trained in our framework are shown.}
\label{tab: benchmark_datasets}
\begin{center}
\resizebox{0.98\columnwidth}{!}{
\begin{tabular}{ l c c c c c}
\toprule
Dataset & Train No. & Test No. & Feature No. & Classifier & Policy network \\
\midrule
MNIST & 60,000 &  10,000 & (1*28*28) & 6-layer CNN & 5-layer CNN \\ 
\midrule
CIFAR-10 & 50,000 & 10,000 &  (3*32*32) & 6-layer CNN & 5-layer CNN \\ 
\midrule
UserTargeting & 19,032 & 19,032 &  153 & 6-layer MLP & 4-, 6-layer MLP \\ 
\bottomrule
\end{tabular}}
\end{center}
\vspace{-2mm}
\end{table}
\endgroup

We verify whether the classifiers learned by our framework are able to yield classification performance improvement on three real world datasets, MNIST, CIFAR-10 and one e-commerce dataset (UserTargeting). Comprehensive experiments are designed to test our proposal with various number of labeled examples and distinct ratios of positive in the unlabeled dataset. Besides, in order to avoid single-sided evaluation, the results of three metrics are presented in the comparison with other state-of-the-art PU learning algorithms.



\subsection{Datasets}

Originally, both MNIST and CIFAR-10 datasets have ten classes. We preprocess and create a binary classification dataset in the same way as \cite{Kiryo2017PositiveUnlabeledLW}. For MNIST, $0, 2, 4, 6, 8$ constitute the P class, while $1, 3, 5, 7, 9$ constitute the N class; For CIFAR-10, {\it airplane,  automobile, ship, truck} form the P class, while {\it bird, cat, deer, dog, frog, horse} are the N class. In our experiments, we first construct PU datasets from their original train dataset, then test the learned classifiers on the original test dataset.

The labeled examples in the UserTargeting dataset are those users who responded positively to certain products on an e-commerce platform. They are used to identify potential users among all other users of this e-commerce platform. We solve this problem by formulating it as a PU learning problem and randomly sample from the users who are not labeled yet as U to create the UserTargeting dataset. 
This dataset is split 50-50 to train and test. Both of them have 4,758 labeled and 14,274 unlabeled users. The feature vector for each user in this dataset is created based on user's past 6 months online activities, and its dimension is 153. The details of the benchmark datasets are shown in Table \ref{tab: benchmark_datasets}.


\subsection{Baselines}
\begin{figure*}
  \centering
  \begin{subfigure}[b]{0.33\linewidth}
  	\includegraphics[width=\linewidth]{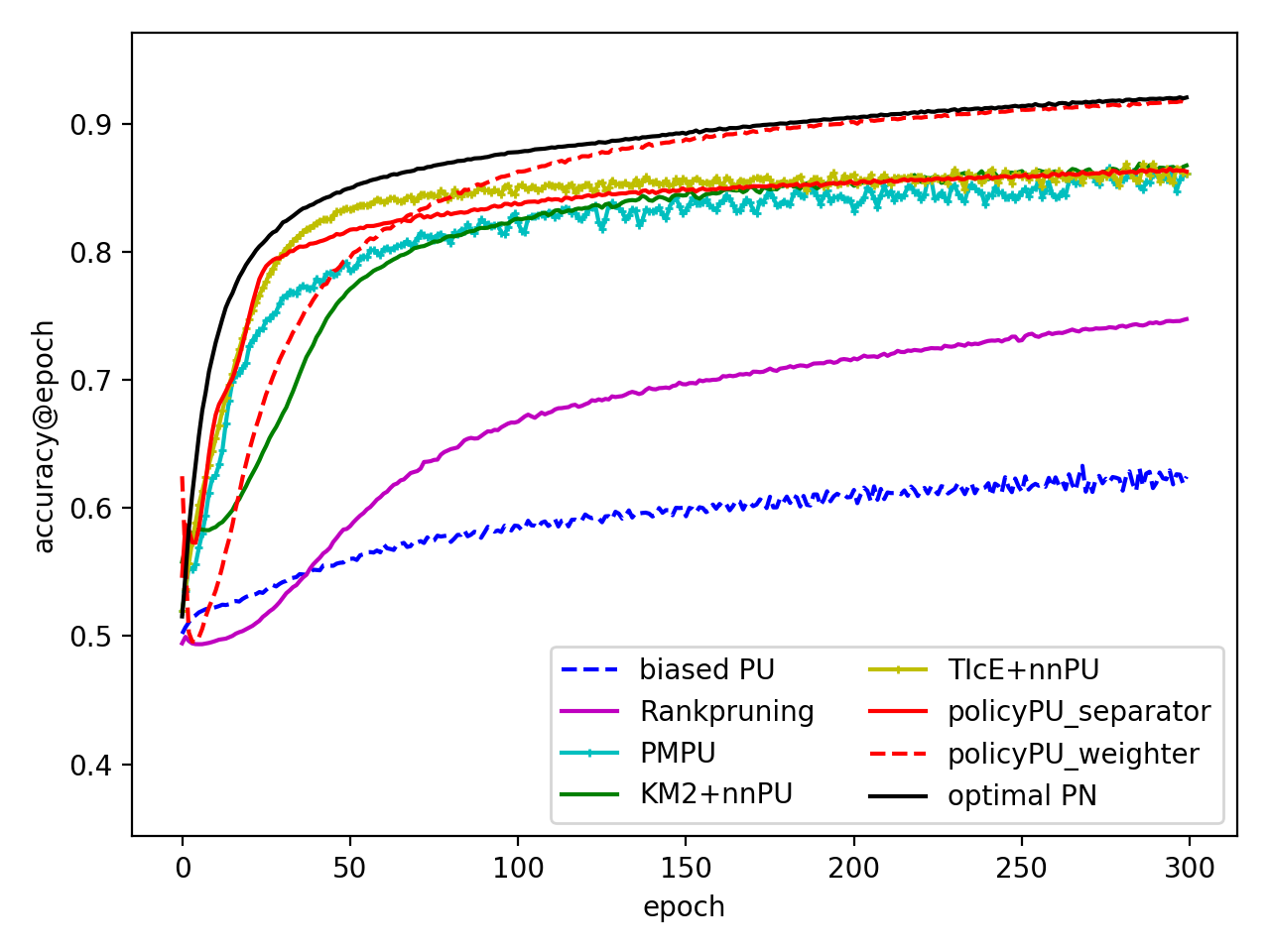}
  \end{subfigure}\hspace*{-0.1em}
  \begin{subfigure}[b]{0.33\linewidth}
  	\includegraphics[width=\linewidth]{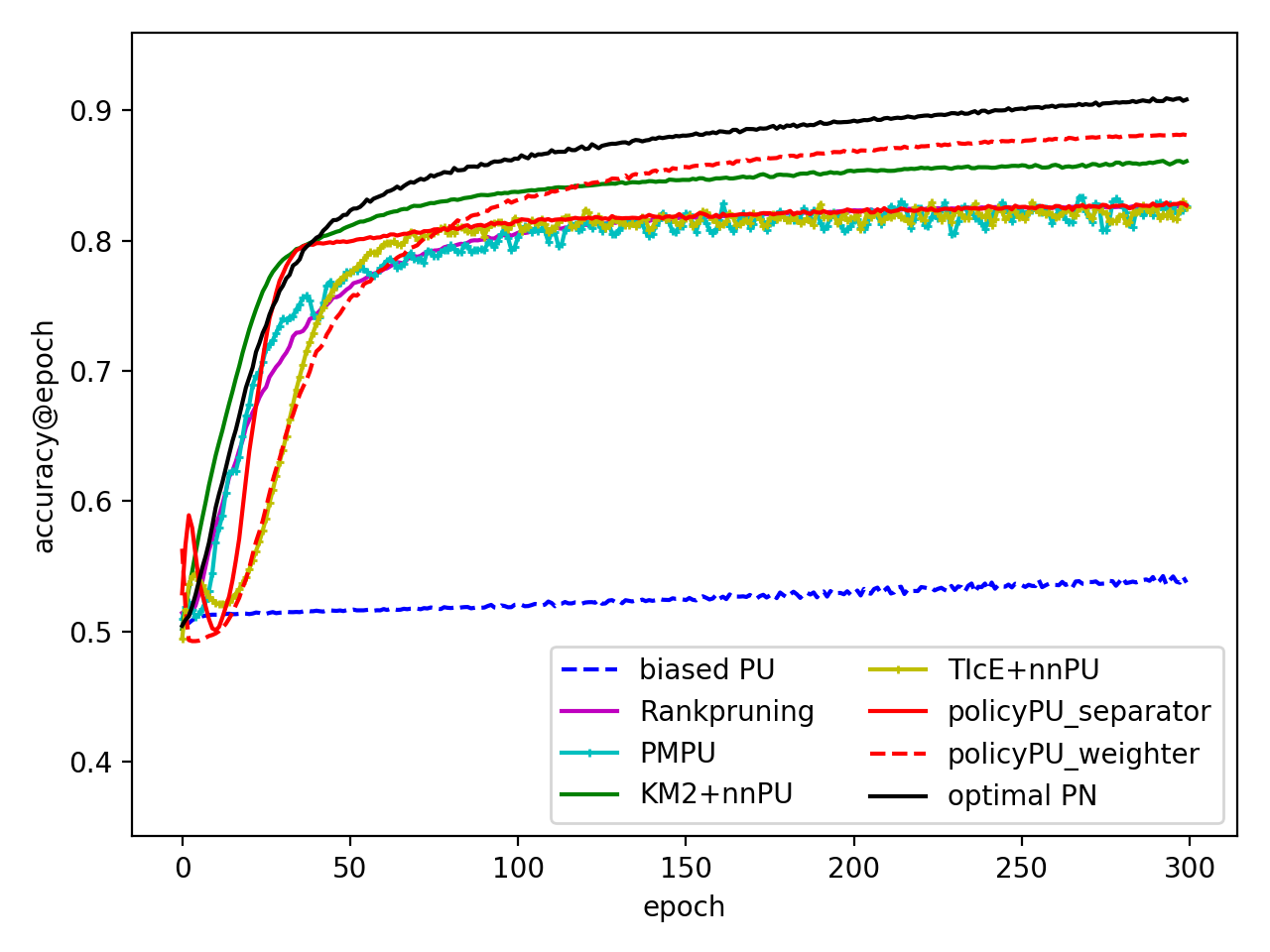}
  \end{subfigure}\hspace*{-0.1em}
  \begin{subfigure}[b]{0.33\linewidth}
  	\includegraphics[width=\linewidth]{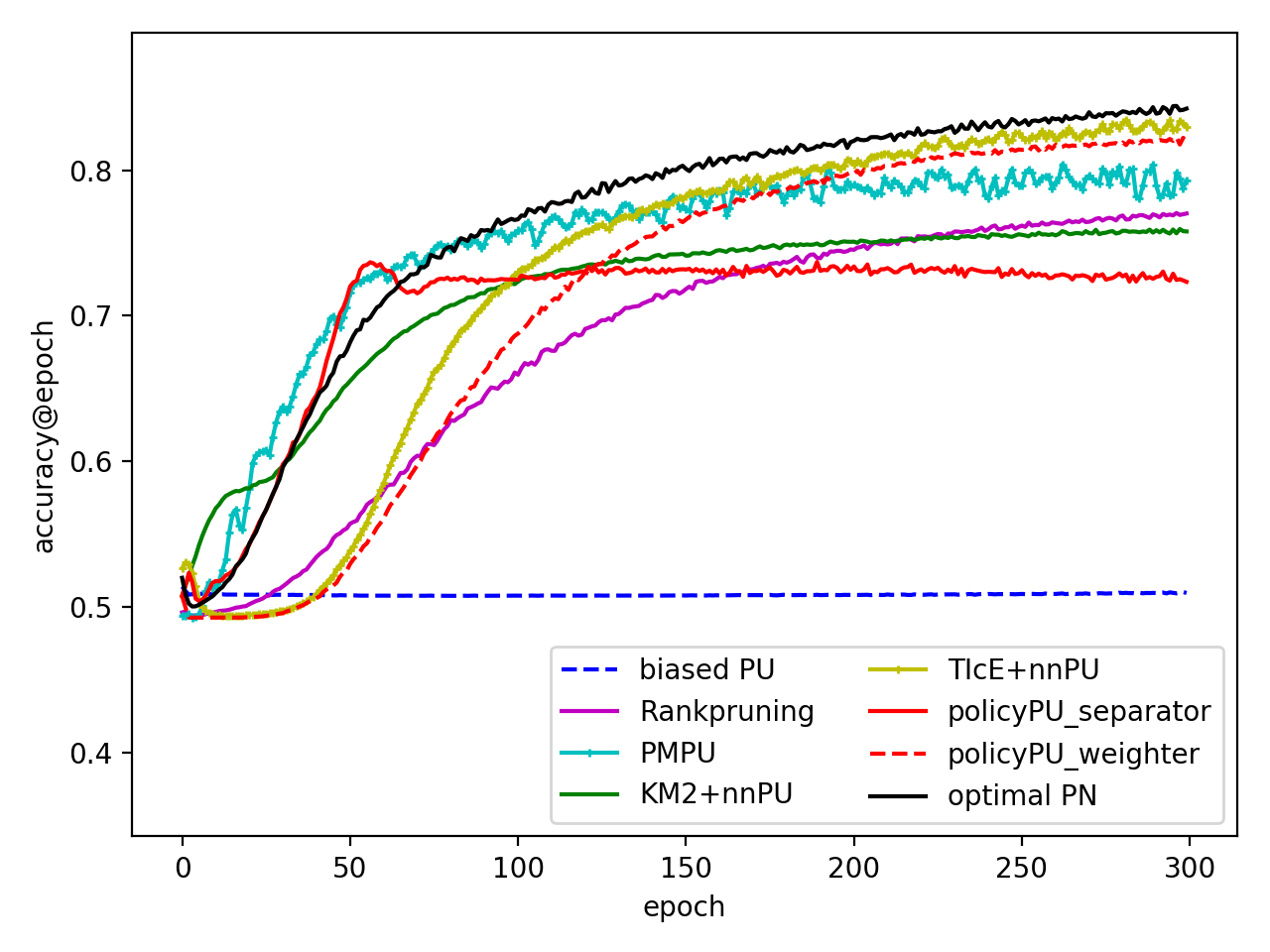}
  \end{subfigure}
  \begin{subfigure}[b]{0.33\linewidth}
  	\includegraphics[width=\linewidth]{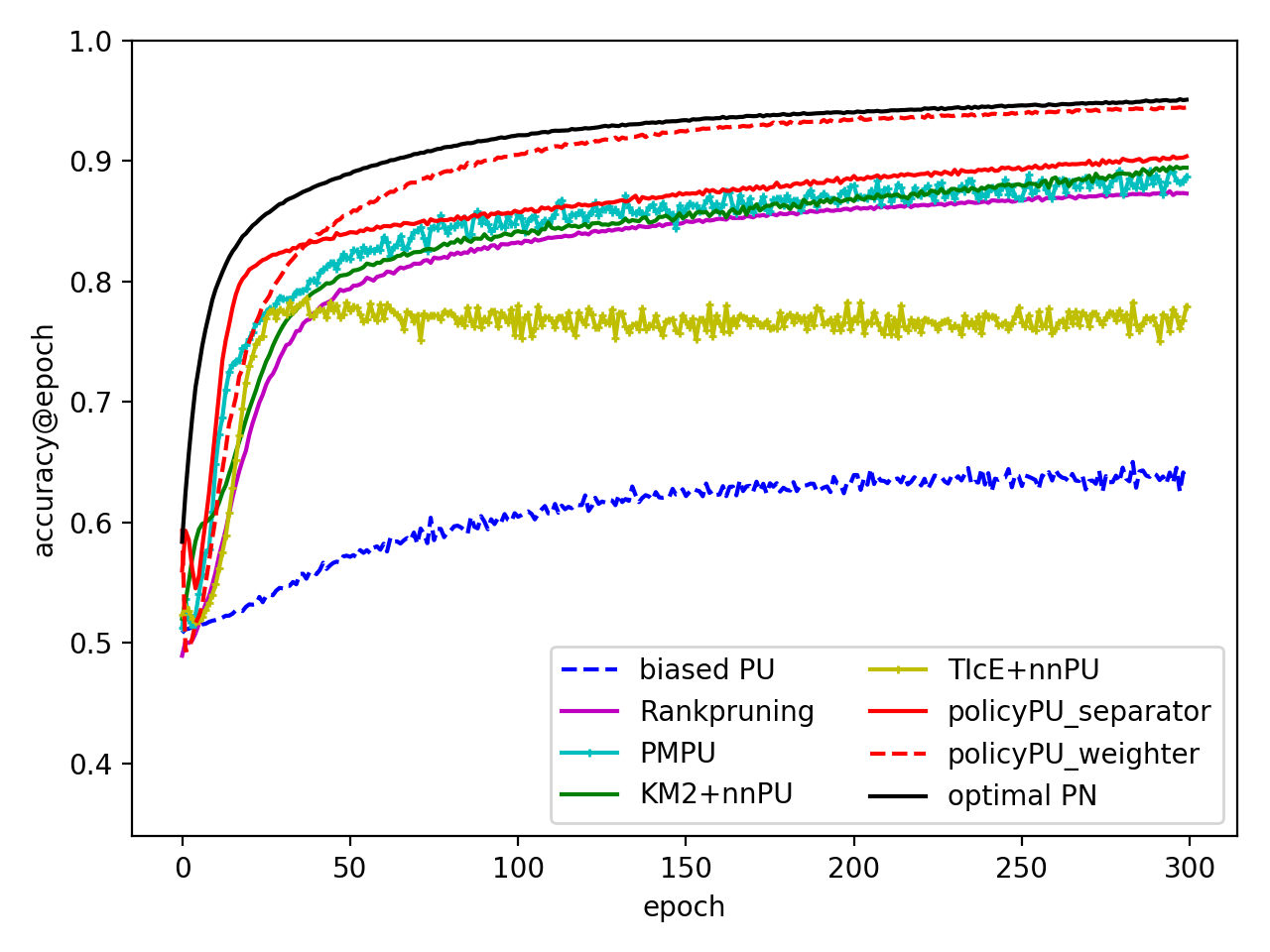}
  \end{subfigure}\hspace*{-0.1em}
  \begin{subfigure}[b]{0.33\linewidth}
  	\includegraphics[width=\linewidth]{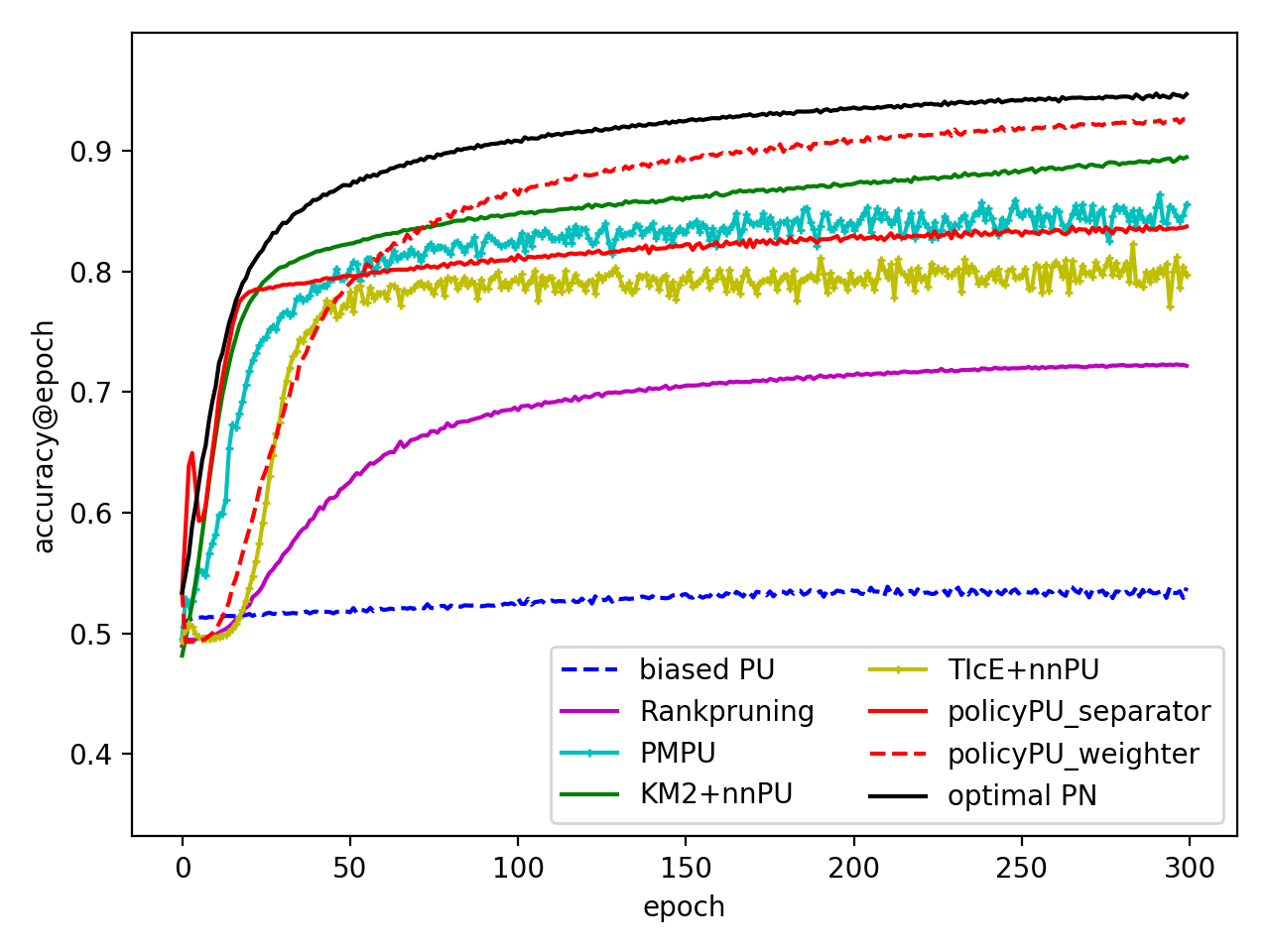}
  \end{subfigure}\hspace*{-0.1em}
  \begin{subfigure}[b]{0.33\linewidth}
  	\includegraphics[width=\linewidth]{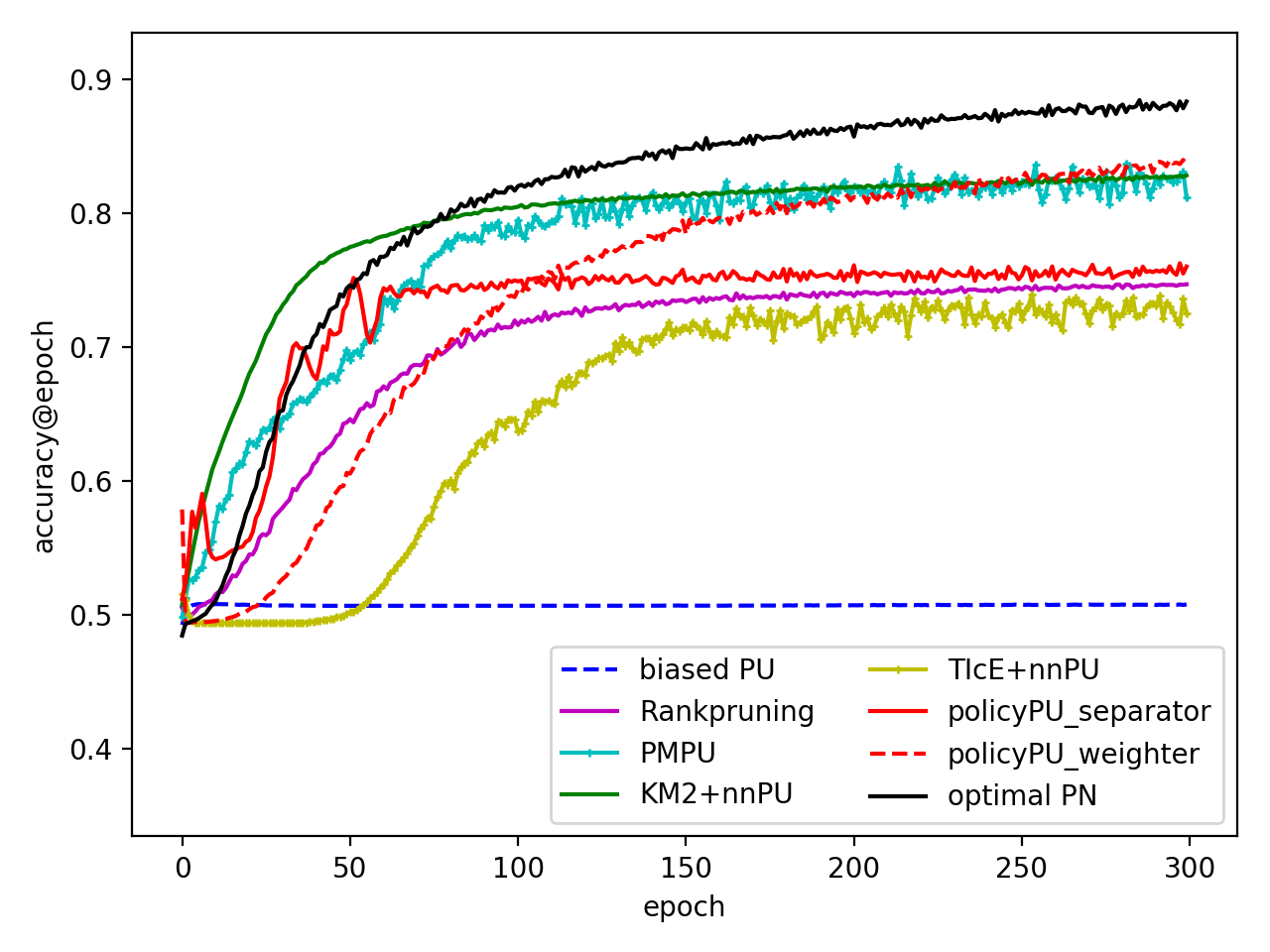}
  \end{subfigure}
  \begin{subfigure}[b]{0.33\linewidth}
  	\includegraphics[width=\linewidth]{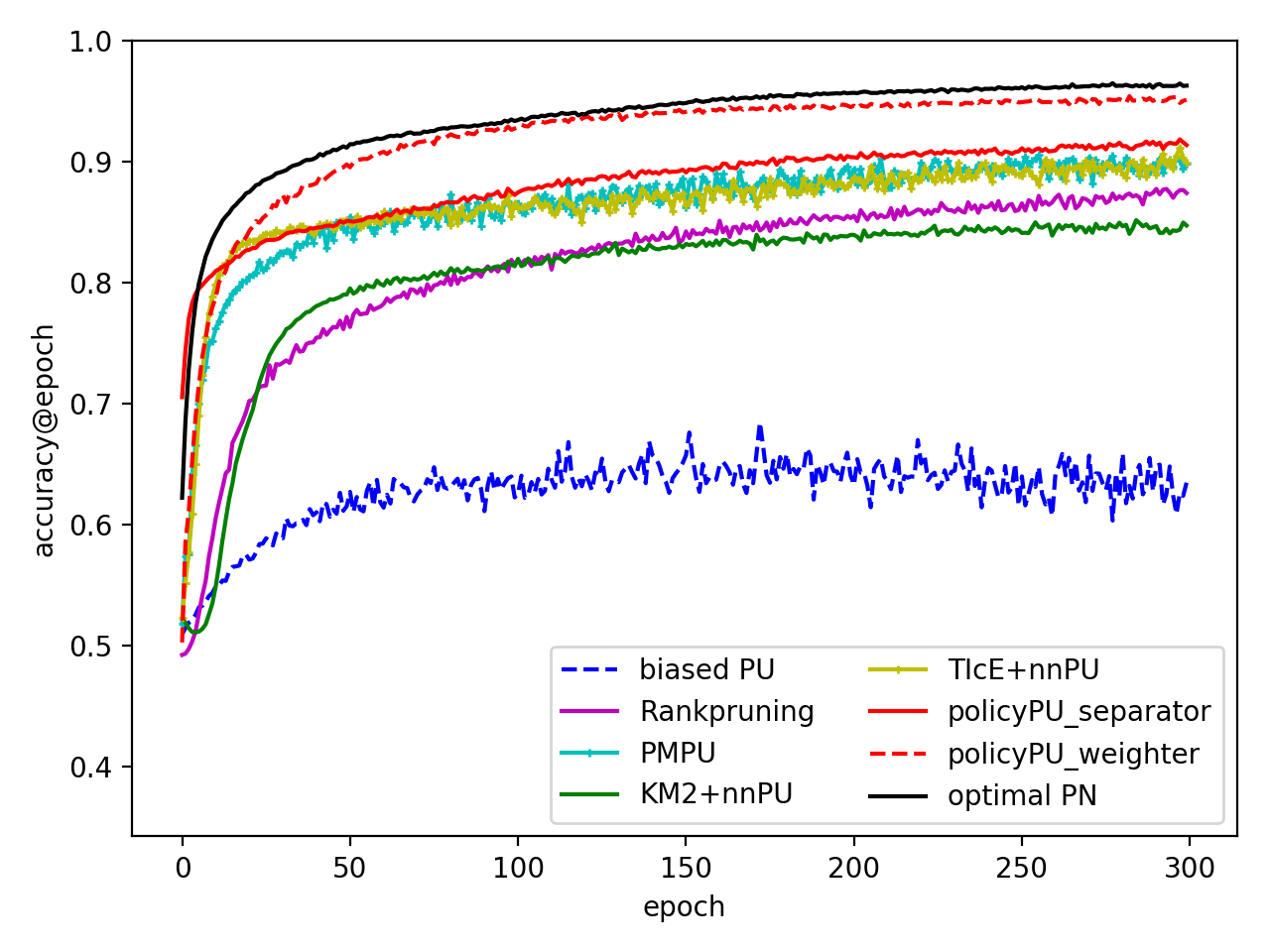}
  \end{subfigure}\hspace*{-0.1em}
  \begin{subfigure}[b]{0.33\linewidth}
  	\includegraphics[width=\linewidth]{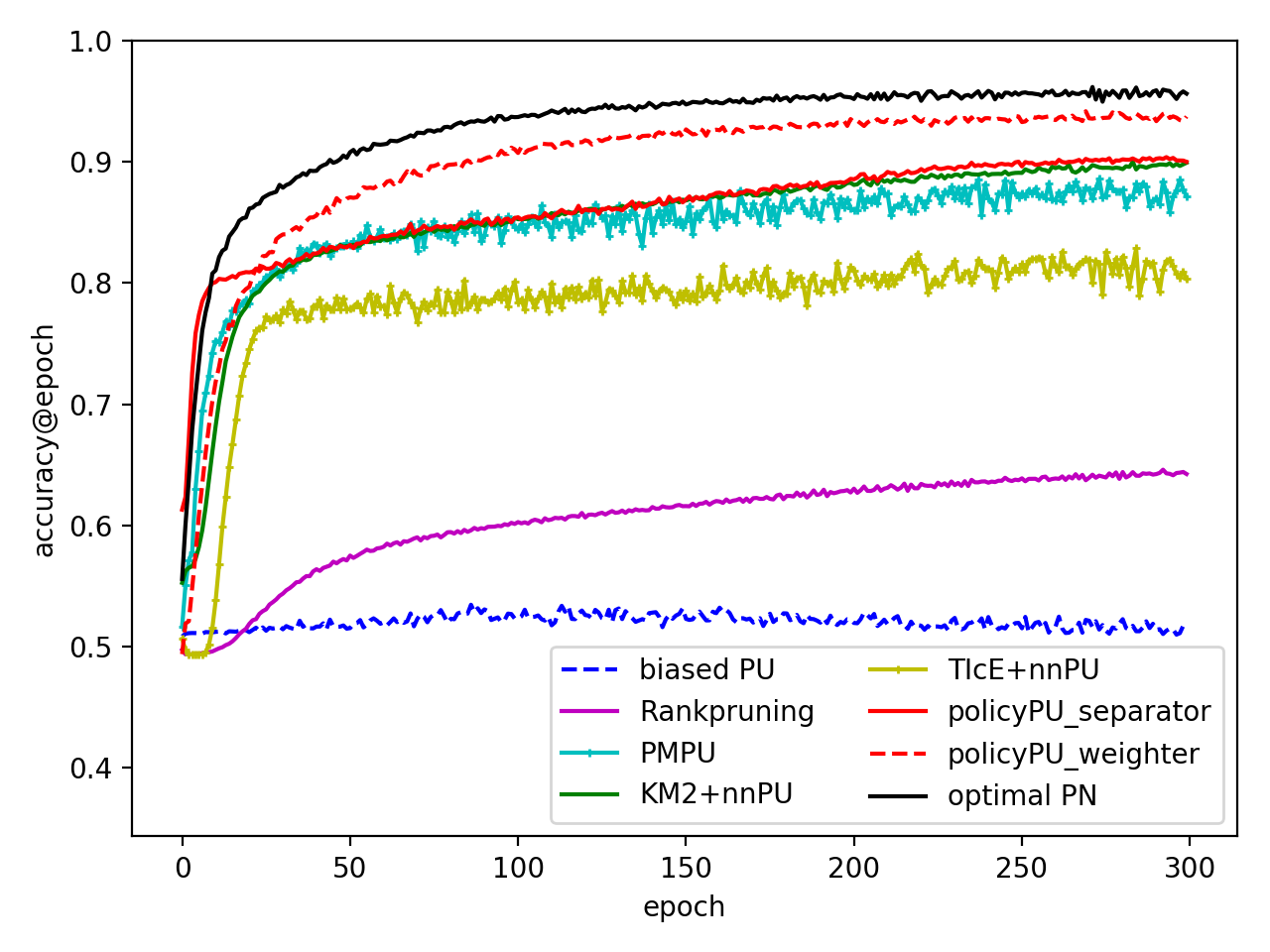}
  \end{subfigure}\hspace*{-0.1em}
  \begin{subfigure}[b]{0.33\linewidth}
  	\includegraphics[width=\linewidth]{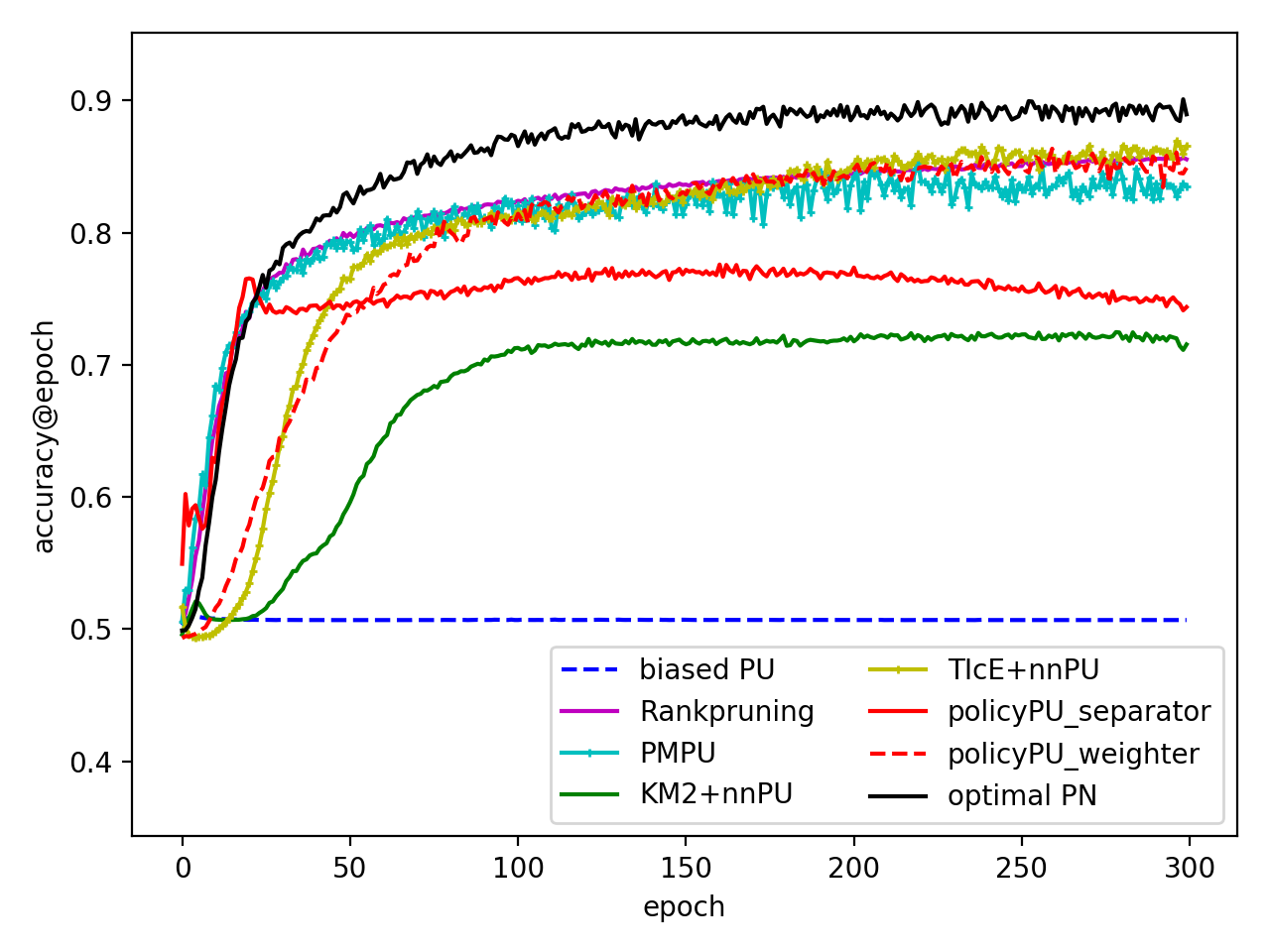}
  \end{subfigure}
  \caption{{\bf Accuracy comparison on {\em MNIST} dataset.} \it The classifier is a 6-layer CNN model, and the policy is a 5-layer CNN model. The number of labeled examples is 300, 500 and 1,000 from the first row to the third row; The percentage of positive examples in the unlabeled data is 0.3, 0.5 and 0.7 from left to right.}
  \label{fig:mnist_cnn}
\end{figure*}
We compare our framework against state-of-the-art PU learning algorithms, which are summarized as follows:

\begin{itemize}
\item biased PU: Biased PU learning builds a classification model by using unlabeled data instances as negatives directly.

\item TIcE\cite{Bekker2018EstimatingTC}+nnPU\cite{Kiryo2017PositiveUnlabeledLW}: TIcE\footnote{\url{https://dtai.cs.kuleuven.be/software/tice/}} utilizes a decision tree induction based method to calculate label frequency first, and obtain class prior via Equation (\ref{eq:alpha_c}). Its estimation result is used as input to nnPU\footnote{\url{https://github.com/kiryor/nnPUlearning}} which is a non-negative unbiased risk estimator.

\item KM2\cite{Ramaswamy2016:MPE}+nnPU\cite{Kiryo2017PositiveUnlabeledLW}: KM2\footnote{\url{http://web.eecs.umich.edu/~cscott/code.html\#kmpe}} is an efficient algorithm for mixture proportion estimation. It embeds the distributions into a reproducing kernel Hilbert space and uses a quadratic programming solver as a sub-routine. The estimation result is also input to nnPU for classifier learning. 

\item RankPruning\cite{northcutt2017:rankpruning}: RankPruning proposes to remove incorrectly labeled examples in the training dataset before inputting them to build classification models. Note that we make an adaptation to its original algorithm\footnote{\url{https://github.com/cgnorthcutt/rankpruning}} that is for $\hat{P}\hat{N}$ learning to fit the PU learning setting.

\item PMPU\cite{Gong2018:MarginBP}: PMPU relies on this large positive margin oracle which claims that positive instances are located far away from the decision boundary. It estimates the labels of unlabeled data in each iteration according to the positive margin shrinkage, and then retrain the classifier based on random sampling.

\item policyPU\_separator: The proposed framework in which a policy network learns to generate a hard assignment of labels to unlabeled examples, while a classifier is built on positives and the output results from the policy. 

\item policyPU\_weighter: The proposed framework that applies a policy network to output continuous actions as the cost function weights for the corresponding classifier.

\item optimal PN: The classification model is built on the training data with ground truth labels. It is used as a reference baseline for other algorithms.
\end{itemize}

ROC$\_$AUC, accuracy and PR$\_$AUC are used as the evaluation metrics.

\subsection{Experiment setup}


We experiment with different numbers of positively labeled examples. Let $n_l$ be the number of labeled examples, $n_l \in \{300, 500, 1000 \}$ are tested for MNIST and CIFAR-10. The number of unlabeled examples in the PU datasets is set to  $3 \times n_l$, and we report the results using varying proportions of positive examples in the unlabeled set, denoted as $\rho$ and set to 0.3, 0.5 and 0.7 for adequate verification.

The targeted classifier for MNIST and CIFAR-10 is a 6-layer CNN with 3 convolutional layers ([d-C(3$\times$3,96)-C(3$\times$3,192)-C(1$\times$1,10)-100-1]), while the policy network is a 5-layer CNN model with 2 convolutional layers([d-C(3$\times$3,96)-C(3$\times$3,10)-100-1]). 
Two 6-layer MLP models are used for the UserTargeting dataset, paired with two different policy network architectures, respectively. The architecture of the classifier and the policy network in our experiments is shown in Table \ref{tab: benchmark_datasets}. 
For policy networks, we deliberately use a slightly shallower architecture compared to the corresponding targeted classifier. The policy is expected to make rough assumptions at the beginning so that it can gradually adjust itself towards the direction of greater cumulative reward.

We train neural networks using $Adam$ as the optimizer with a batch size of 128 and a learning rate fixed to $1e\mbox{-}5$. 
Also, we use $ReLU$ \cite{Maas13:rectifiernonlinearities} as the activation function, apply weight decay and batch normalization\cite{Ioffe2015:bn}. To achieve fair comparison, we train classifiers with same architecture and same parameter settings for different algorithms on all created PU datasets.  We average the performance by running each experiment 5 times. 
All experiments are implemented using Chainer~\footnote{https://chainer.org/}.

\subsection{Experiments on the MNIST dataset}

In the experiment, a PU dataset is first created for each setting with respect to the number of labeled examples and the positive ratio in the unlabeled data.
For TIcE and KM2, the feature vectors are first flattened, and then fed to generate class priors. We follow their default settings, and conduct downsampling if the total number of training instances exceeds $2,000$, the same as \cite{Bekker2018EstimatingTC}, for better estimation. 
Then, these estimated values are input to nnPU for training classification models.
The classifier in RankPruning is first trained to remove noisy labels in the unlabeled dataset. We follow its setting to run a 5-folder cross validation in order to prune incorrectly labeled examples and get weights for different classes. Then, its classification model is learned with a weighted cost function on the pruned dataset.
For PMPU, we adapt it to a mini-batch training scenario, obtain the large positive margin oracle $\tau$ and resample 3/4 unlabeled examples for classifier learning within every mini-batch.
As for the optimal PN, the classification model is built with both labeled positive and negative examples. The performance of optimal PN is used as a reference to other PU learning algorithms.
The weight decay for all classifiers is set to $2.0$, while that of the policy network in \emph{Weighter} is $2.0$ and in \emph{Separator} is $0.5$. We pre-train the policy networks and classifiers using unlabeled examples as negatives for 5 epochs. Then, we run $300$ epochs for training and update policy every $3$ epochs. 

The accuracy with different settings is displayed in Fig. \ref{fig:mnist_cnn}. 
Vertically, it shows experimental results with different number of labeled examples, 300, 500 and 1,000 from top row to the bottom. Horizontally, the figures are different in terms of the percentage of P in U. 
The experimental results verify that the classifiers learned by the proposed framework are very competitive with different number of labeled examples in training datasets, as well as the different ratios of positive examples in the unlabeled data. 

Especially, when the ratio of positive data is low (\ie, $\rho$=0.3), policyPU\_weighter is shown to be capable of approaching to the accuracy curve of optimal PN learning. 
We conjecture the reason is that the data used for the classifier to obtain the \emph{threshold} have a relatively large proportion of labeled true positive data. As a consequence, it is able to produce valid reward to those unlabeled data.
Meanwhile, another observation is that with a larger $\rho$, it takes longer for both of the classifiers in our framework to start to predict reasonably.
Particularly, policyPU\_separator is hardly able to keep increasing its performance when $\rho=0.7$, likely due to the influence of the \emph{threshold} setting in Equation (\ref{eq:return_func}). As described, the threshold is calculated based on the expectation over the predicted class label probability of labeled examples and some unlabeled examples chosen based on $thresh_{min}$. Compared to solely using labeled examples as the reference, the proposed way is expected to get a balanced threshold value by considering those unlabeled examples which are likely to be positive. Yet, it is also possible to increase the threshold if many positives are far from negatives in the unlabeled dataset. As a result, the policy may get non-optimal reward from those positives in U data near the decision boundary due to the threshold setting. 
On the other hand, the policyPU\_weighter is not impacted as severe as the policyPU\_separator. We think it is because of the weighting mechanism that the classifier in \emph{Weighter} holds. It is capable of drawing a more flexible decision boundary even with many data instances near it. Therefore, the class label prediction would be more accurate and more valid as reward to the policy.

The performance comparison after 300 epochs training is shown in Table \ref{tab: mnist_300}, \ref{tab: mnist_500} and \ref{tab: mnist_1000} for the experiments with 300, 500 and 1,000 labeled examples, respectively. Each table shows the ROC\_AUC, accuracy and PR\_AUC results of three distinct fraction settings. It is shown that almost all algorithms can generate consistent performance except biased PU learning which fails to achieve a good accuracy. 
Experiment results show that the classifiers learned by the proposed framework can outperform others and even output close results to optimal results for a few cases.



\begin{table}[t!]
\caption{{\bf Experiment results on {\em MNIST} with CNN classifiers.} \it No. of labeled examples is 300. The percentage of P in U is 0.3, 0.5 and 0.7. ROC\_AUC, accuracy and PR\_AUC are shown from left to right for each percentage setting.
}
\begin{center}
\label{tab: mnist_300} 
\resizebox{\columnwidth}{!}{
\begin{tabular}{ l c  c c c |c c c| c c c }
\toprule

Model && \multicolumn{3}{c}{{\em 0.3}} & \multicolumn{3}{c}{{\em 0.5}} & \multicolumn{3}{c}{{\em 0.7}} \\
\cmidrule{1-2} \cmidrule{3-5} \cmidrule{6-8} \cmidrule{9-11} 

biased PU && 0.957 & 0.622 & 0.957  &  0.929 & 0.535 & 0.933 &  0.874 & 0.510 & 0.865 \\

TIcE+nnPU && 0.953 & 0.860 & 0.954  &  0.936 & 0.825 & 0.936 &   \bftab0.942 &\bftab0.828& \bftab0.940  \\

KM2+nnPU && 0.951 & 0.867  & 0.954  &  0.938 & 0.861 & 0.943  &   0.906 &0.722& 0.905 \\

RankPruning && 0.875 & 0.745 & 0.859   &   0.899  & 0.825 & 0.882   &   0.878 &0.771& 0.874 \\

PMPU && 0.956 & 0.861 & 0.957   &  0.931 & 0.827 & 0.933   &   0.911 &0.793& 0.911 \\
\cmidrule{1-2} \cmidrule{3-5} \cmidrule{6-8} \cmidrule{9-11}

policyPU\_separator && 0.954 & 0.862 & 0.957   &   0.927 & 0.825 & 0.934   &   0.890 &0.722& 0.893 \\

policyPU\_weighter && \bftab0.975 & \bftab0.916 & \bftab0.975   &   \bftab0.948 & \bftab0.880 & \bftab0.948   &   0.915 &0.818& 0.909 \\
\cmidrule{1-2} \cmidrule{3-5} \cmidrule{6-8} \cmidrule{9-11}

optimal PN && 0.976 & 0.919 & 0.977  &  0.974 & 0.907 & 0.973   & 0.971 &0.839 &  0.969   \\
\bottomrule
\end{tabular}}
\end{center}
\end{table}

\begin{table}[t!]
\caption{{\bf Experiment results on {\em MNIST} with CNN classifiers.} \it No. of labeled examples is 500. The percentage of P in U is 0.3, 0.5 and 0.7. ROC\_AUC, accuracy and PR\_AUC are shown from left to right for each percentage setting.
}
\begin{center}
\label{tab: mnist_500} 
\resizebox{\columnwidth}{!}{
\begin{tabular}{ l c  c c c |c c c| c c c }
\toprule
Model && \multicolumn{3}{c}{{\em 0.3}} & \multicolumn{3}{c}{{\em 0.5}} & \multicolumn{3}{c}{{\em 0.7}} \\
\cmidrule{1-2} \cmidrule{3-5} \cmidrule{6-8} \cmidrule{9-11} 

biased PU && 0.976 & 0.632 & 0.976  &  0.955 & 0.536 & 0.955   &  0.898 &0.508& 0.902 \\

TIcE+nnPU && 0.963 & 0.778 & 0.962  &  0.962 & 0.794 & 0.960   & 0.948 &0.722& 0.944  \\

KM2+nnPU && 0.970 & 0.892 & 0.971  &  0.962 & 0.894 & 0.963   &   0.924 &0.828& 0.934 \\

RankPruning && 0.941 & 0.872 & 0.924   &   0.770 & 0.722 & 0.722   &   0.832 &0.746& 0.820 \\

PMPU && 0.971 & 0.885 & 0.972   &   0.954 & 0.856 & 0.954   &   0.921 &0.811& 0.929 \\
\cmidrule{1-2} \cmidrule{3-5} \cmidrule{6-8} \cmidrule{9-11}

policyPU\_separator && 0.973 & 0.903 & 0.975   &   0.943 & 0.836 & 0.947   &   0.903 &0.760& 0.917 \\

policyPU\_weighter && \bftab0.986 & \bftab0.942 & \bftab0.986    &  \bftab 0.979 & \bftab0.923 & \bftab0.979   &   \bftab0.950 &\bftab0.833& \bftab0.950 \\
\cmidrule{1-2} \cmidrule{3-5} \cmidrule{6-8} \cmidrule{9-11}

optimal PN && 0.989 & 0.949 & 0.989  &  0.990 & 0.944 & 0.990  & 0.986 &0.877 &  0.985   \\
\bottomrule
\end{tabular}}
\end{center}
\end{table}

\begin{table}[t!]
\caption{{\bf Experiment results on {\em MNIST} with CNN classifiers.} \it No. of labeled examples is 1,000. The percentage of P in U is 0.3, 0.5 and 0.7. ROC\_AUC, accuracy and PR\_AUC are shown from left to right for each percentage setting.
}
\begin{center}
\label{tab: mnist_1000} 
\resizebox{\columnwidth}{!}{
\begin{tabular}{ l c  c c c |c c c| c c c }
\toprule
Model && \multicolumn{3}{c}{{\em 0.3}} & \multicolumn{3}{c}{{\em 0.5}} & \multicolumn{3}{c}{{\em 0.7}} \\
\cmidrule{1-2} \cmidrule{3-5} \cmidrule{6-8} \cmidrule{9-11} 

biased PU && 0.983 & 0.626 & 0.983  &  0.973 & 0.517 & 0.973   &  0.937 &0.507& 0.946 \\

TIcE+nnPU && 0.979 & 0.893 & 0.978  &  0.974 & 0.794 & 0.972   & 0.975 &\bftab0.858& 0.975  \\

KM2+nnPU && 0.969 & 0.845 & 0.971  &  0.973 & 0.897 & 0.974   &   0.905 &0.715& 0.917 \\

RankPruning && 0.965 & 0.869 & 0.963   &   0.776 & 0.639 & 0.763   &   0.932 &0.853& 0.932 \\

PMPU && 0.980 & 0.895 & 0.979   &   0.970 & 0.871 & 0.970   &   0.931 &0.833& 0.938 \\
\cmidrule{1-2} \cmidrule{3-5} \cmidrule{6-8} \cmidrule{9-11}

policyPU\_separator && 0.979 & 0.912 & 0.979   &   0.969 & 0.898 & 0.971   &   0.879 &0.743& 0.896 \\

policyPU\_weighter && \bftab0.991 & \bftab0.948 & \bftab0.991    &  \bftab 0.989 & \bftab0.935 & \bftab0.988   &   \bftab0.978 &0.843& \bftab0.977 \\
\cmidrule{1-2} \cmidrule{3-5} \cmidrule{6-8} \cmidrule{9-11}

optimal PN && 0.993 & 0.960 & 0.993  &  0.994 & 0.952 & 0.993  &  0.991 &0.884 &  0.990   \\
\bottomrule
\end{tabular}}
\end{center}
\vspace{-2mm}
\end{table}

\subsection{Experiments on the CIFAR-10 dataset}

\begin{figure*}
  \centering
  
  \begin{subfigure}[b]{0.33\linewidth}
  	\includegraphics[width=\linewidth]{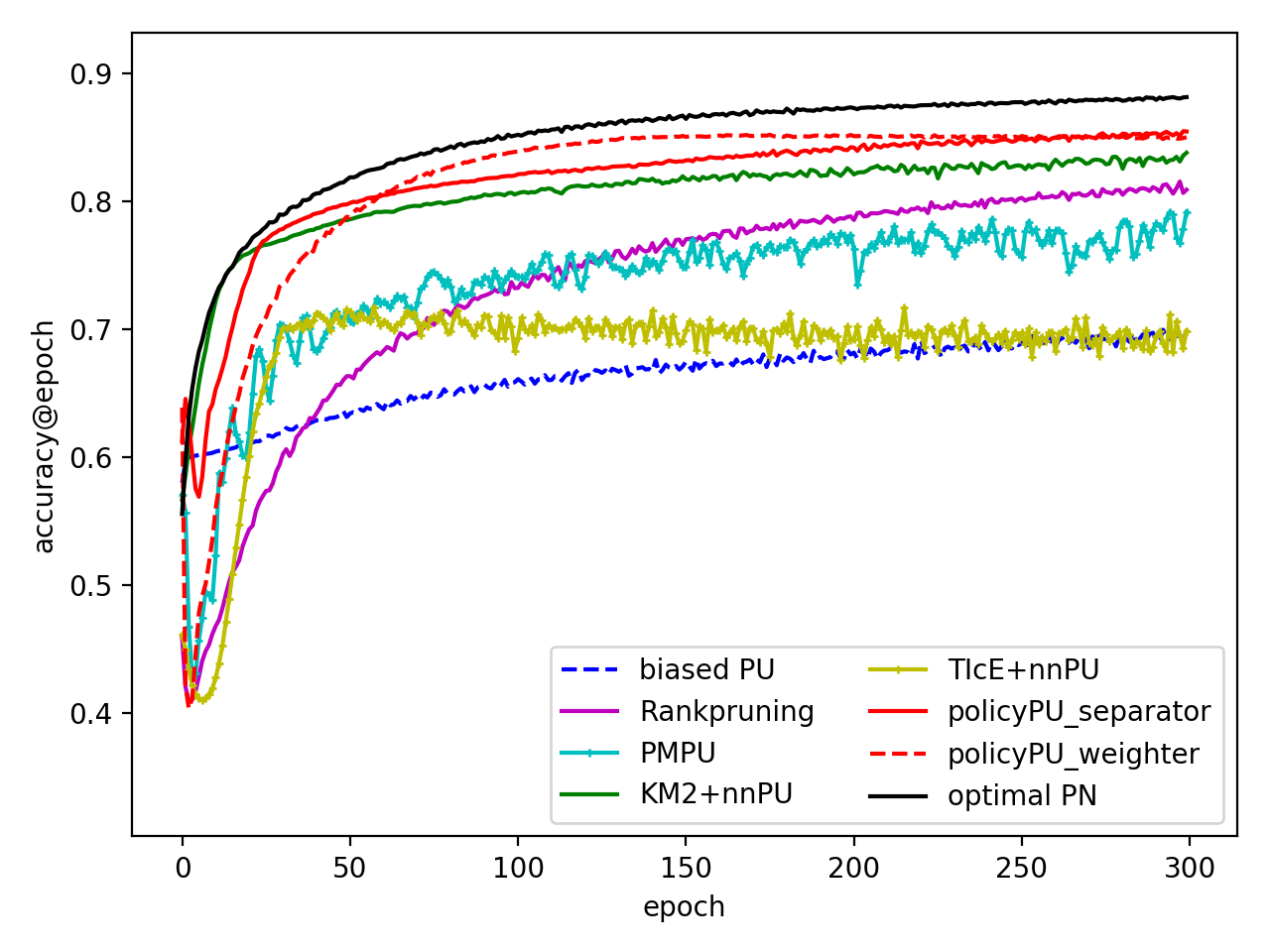}
  \end{subfigure}\hspace*{-0.1em}
  \begin{subfigure}[b]{0.33\linewidth}
  	\includegraphics[width=\linewidth]{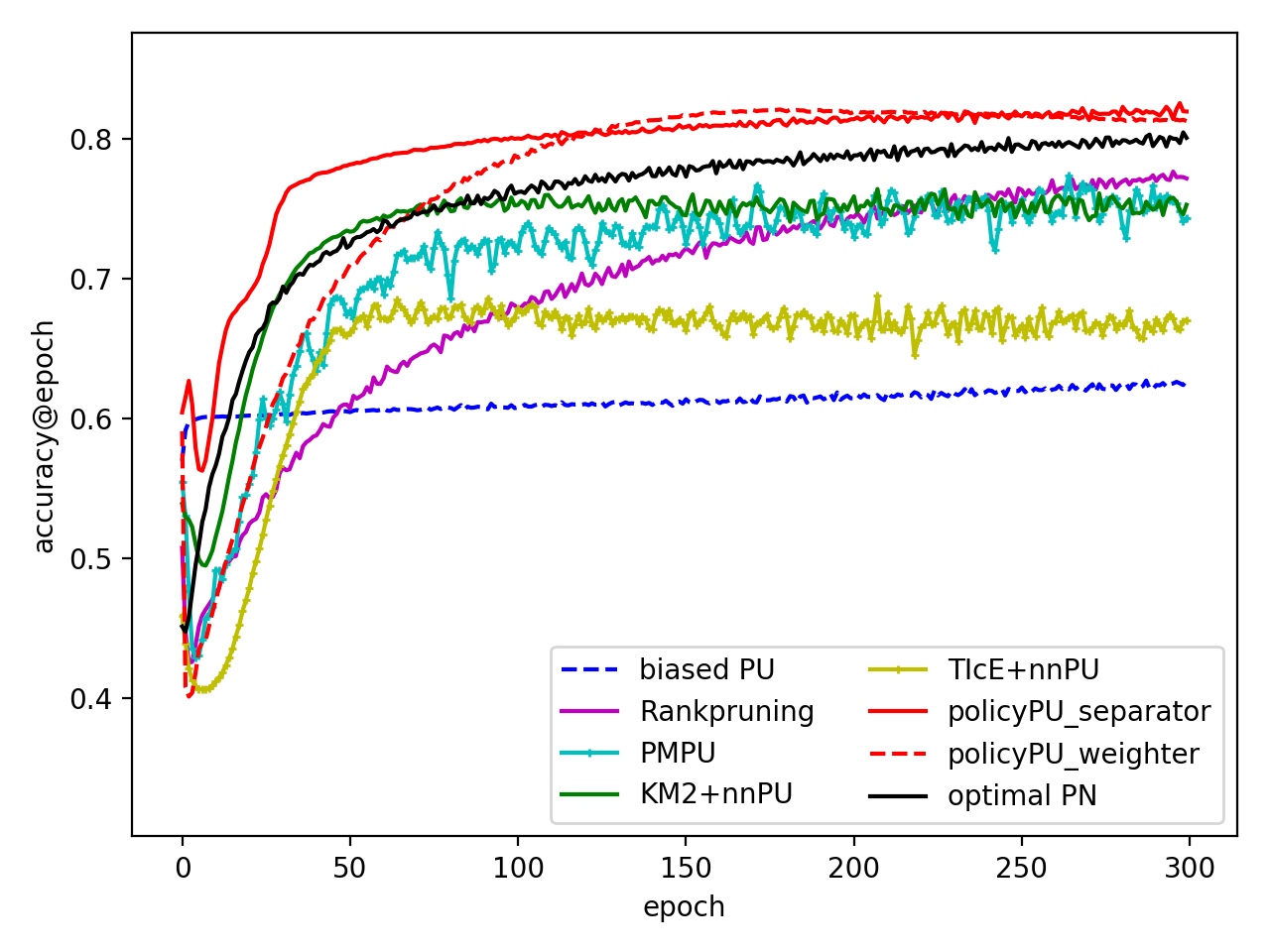}
  \end{subfigure}\hspace*{-0.1em}
  \begin{subfigure}[b]{0.33\linewidth}
  	\includegraphics[width=\linewidth]{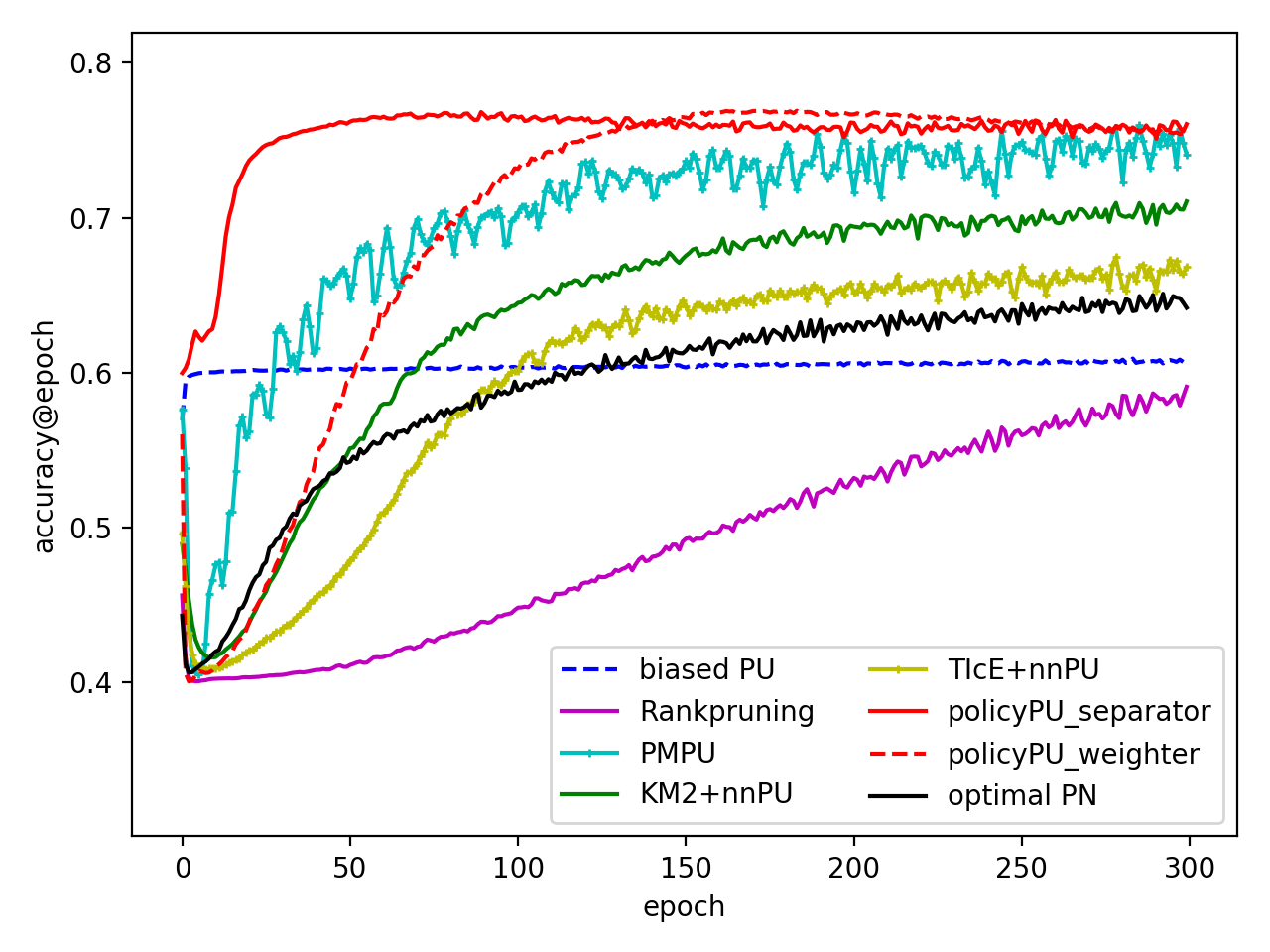}
  \end{subfigure}
  \begin{subfigure}[b]{0.33\linewidth}
  	\includegraphics[width=\linewidth]{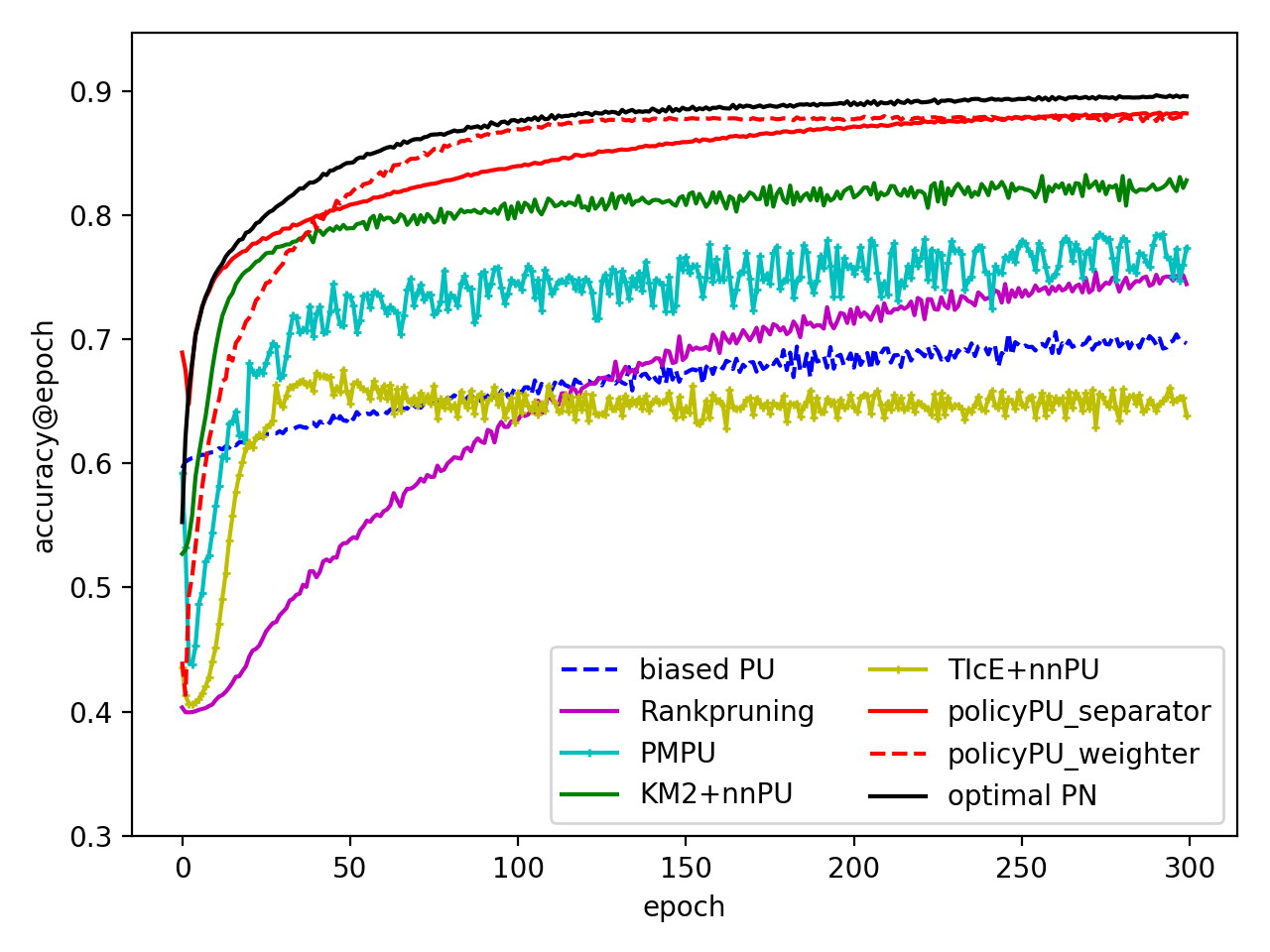}
  \end{subfigure}\hspace*{-0.1em}
  \begin{subfigure}[b]{0.33\linewidth}
  	\includegraphics[width=\linewidth]{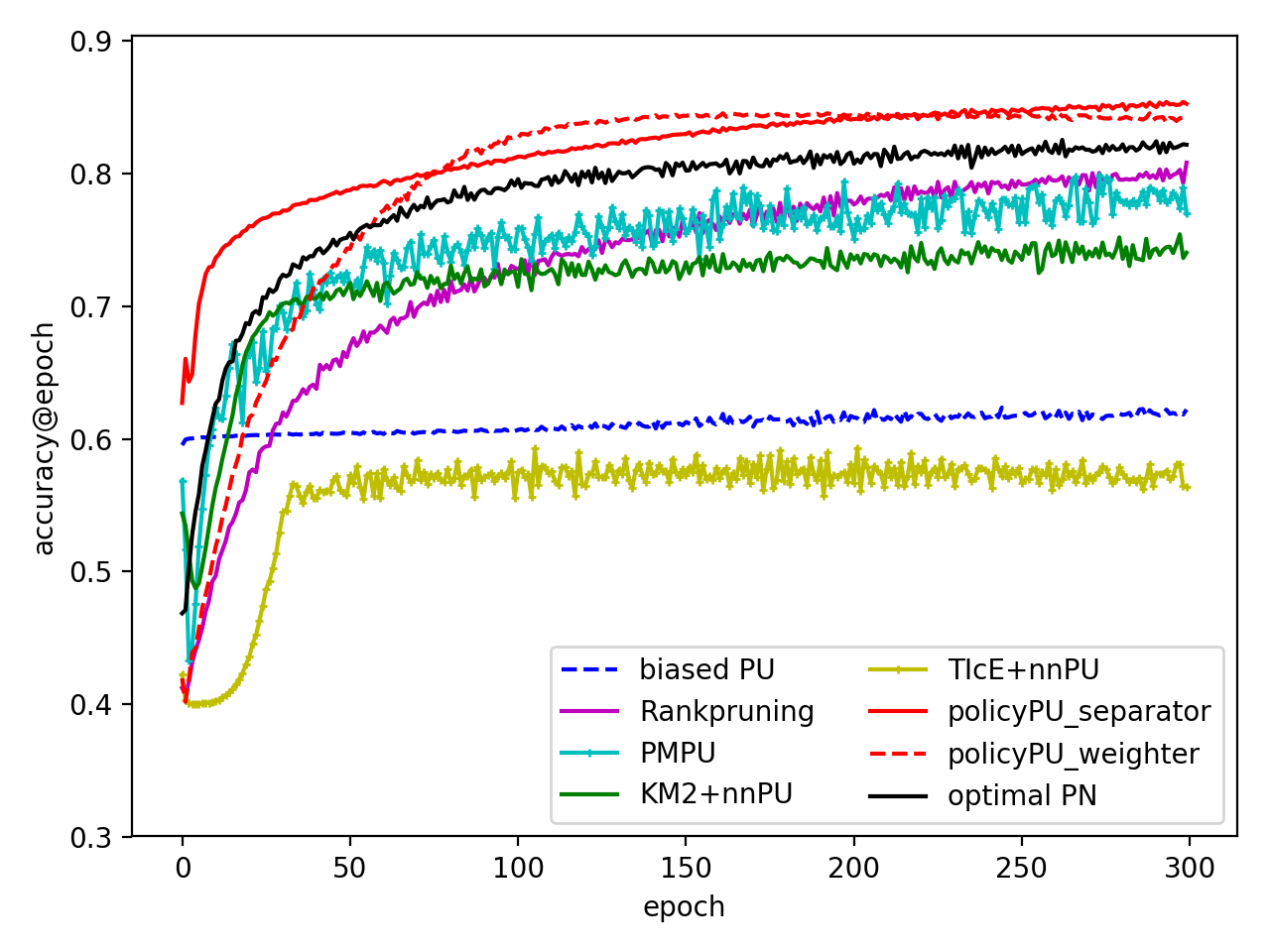}
  \end{subfigure}\hspace*{-0.1em}
  \begin{subfigure}[b]{0.33\linewidth}
  	\includegraphics[width=\linewidth]{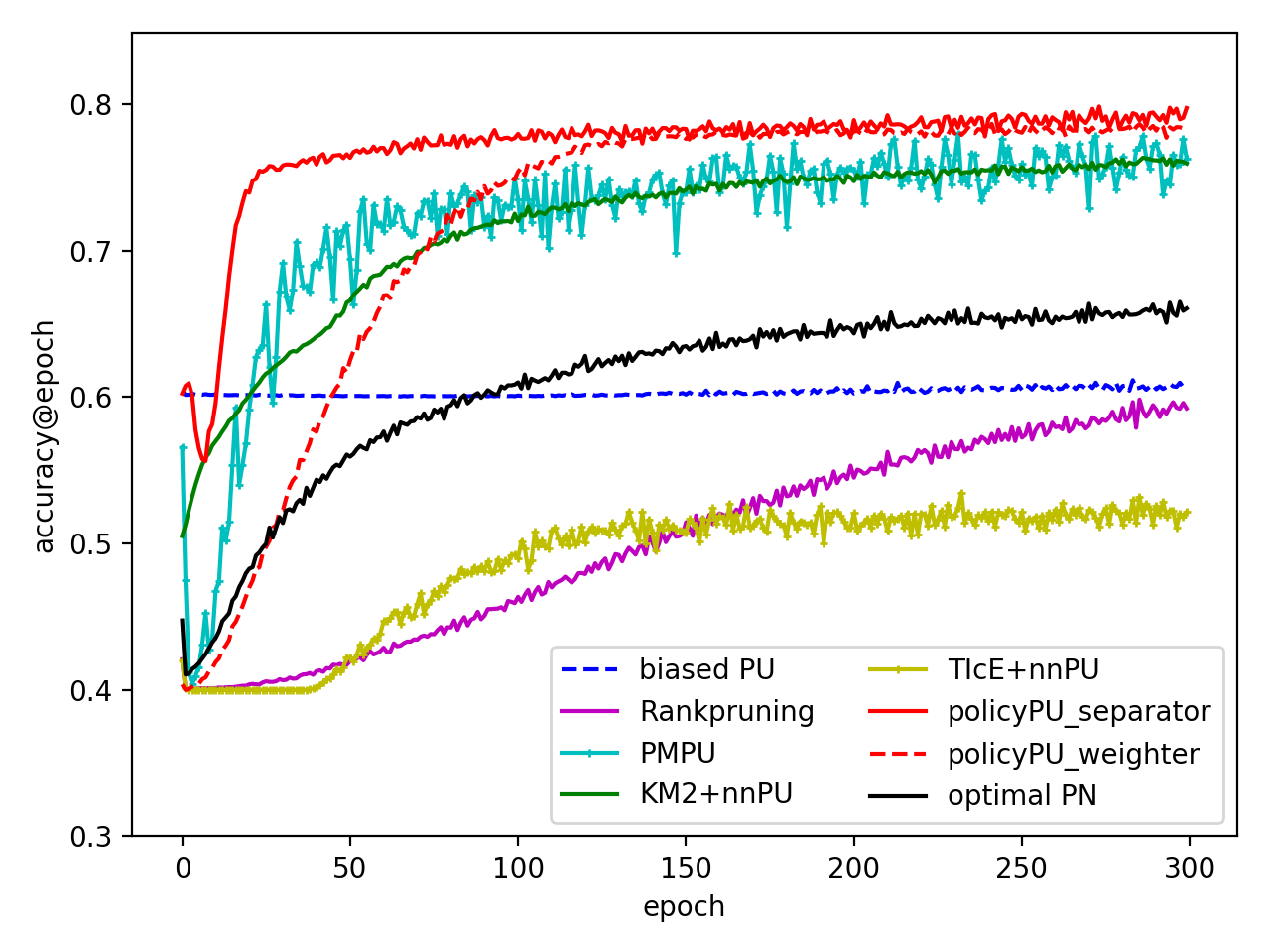}
  \end{subfigure}
  \begin{subfigure}[b]{0.33\linewidth}
  	\includegraphics[width=\linewidth]{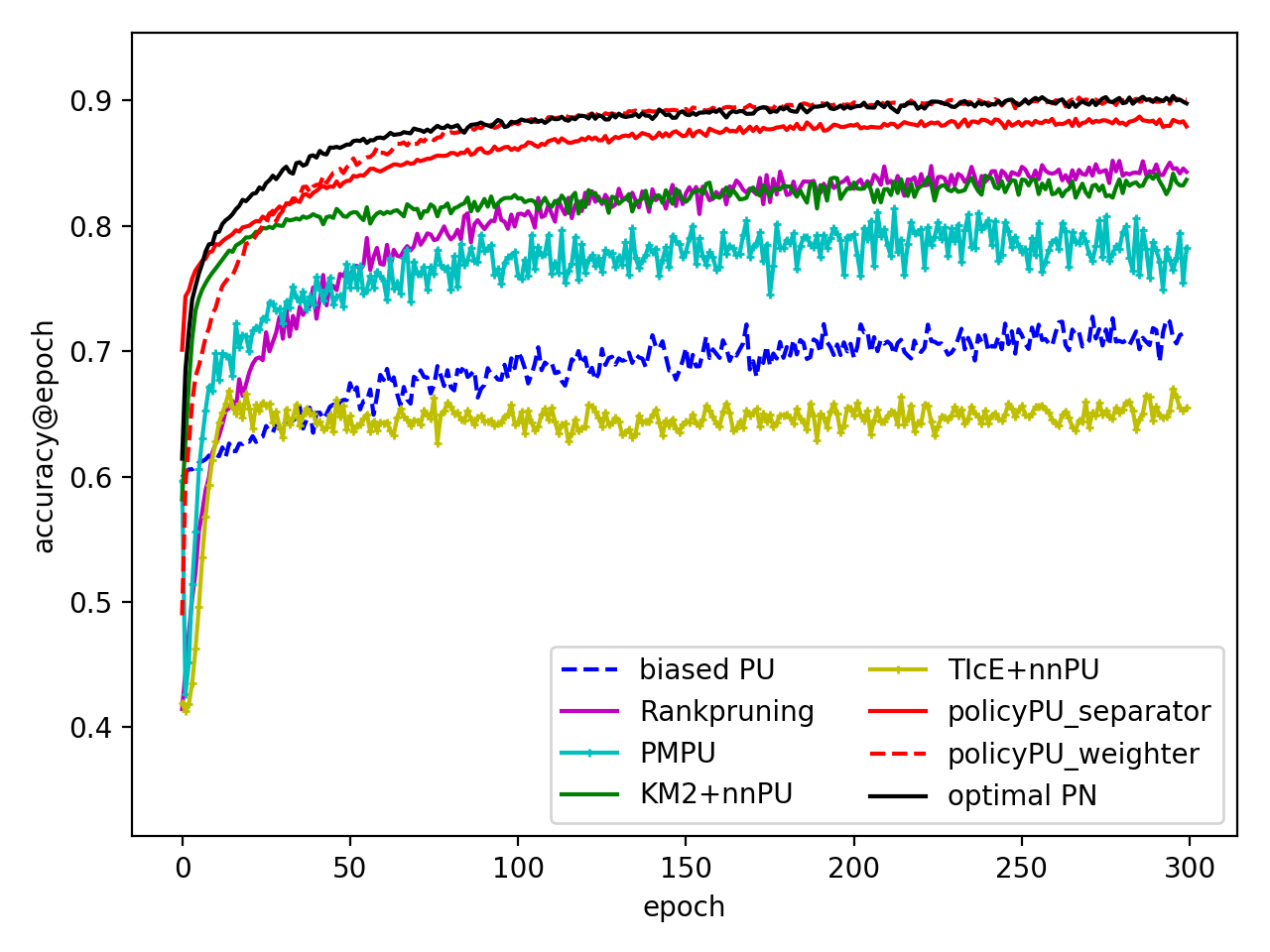}
  \end{subfigure}\hspace*{-0.1em}
  \begin{subfigure}[b]{0.33\linewidth}
  	\includegraphics[width=\linewidth]{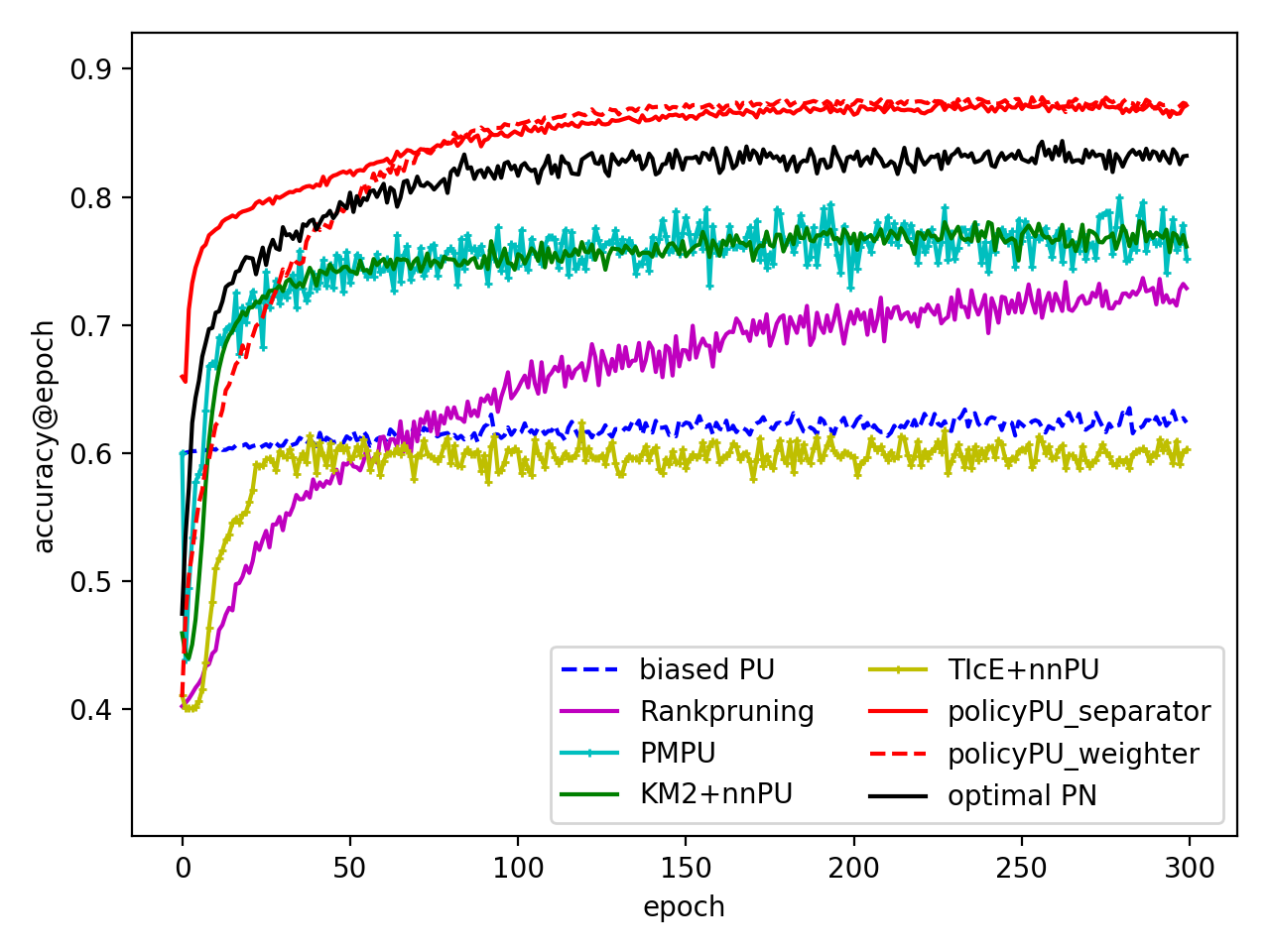}
  \end{subfigure}\hspace*{-0.1em}
  \begin{subfigure}[b]{0.33\linewidth}
  	\includegraphics[width=\linewidth]{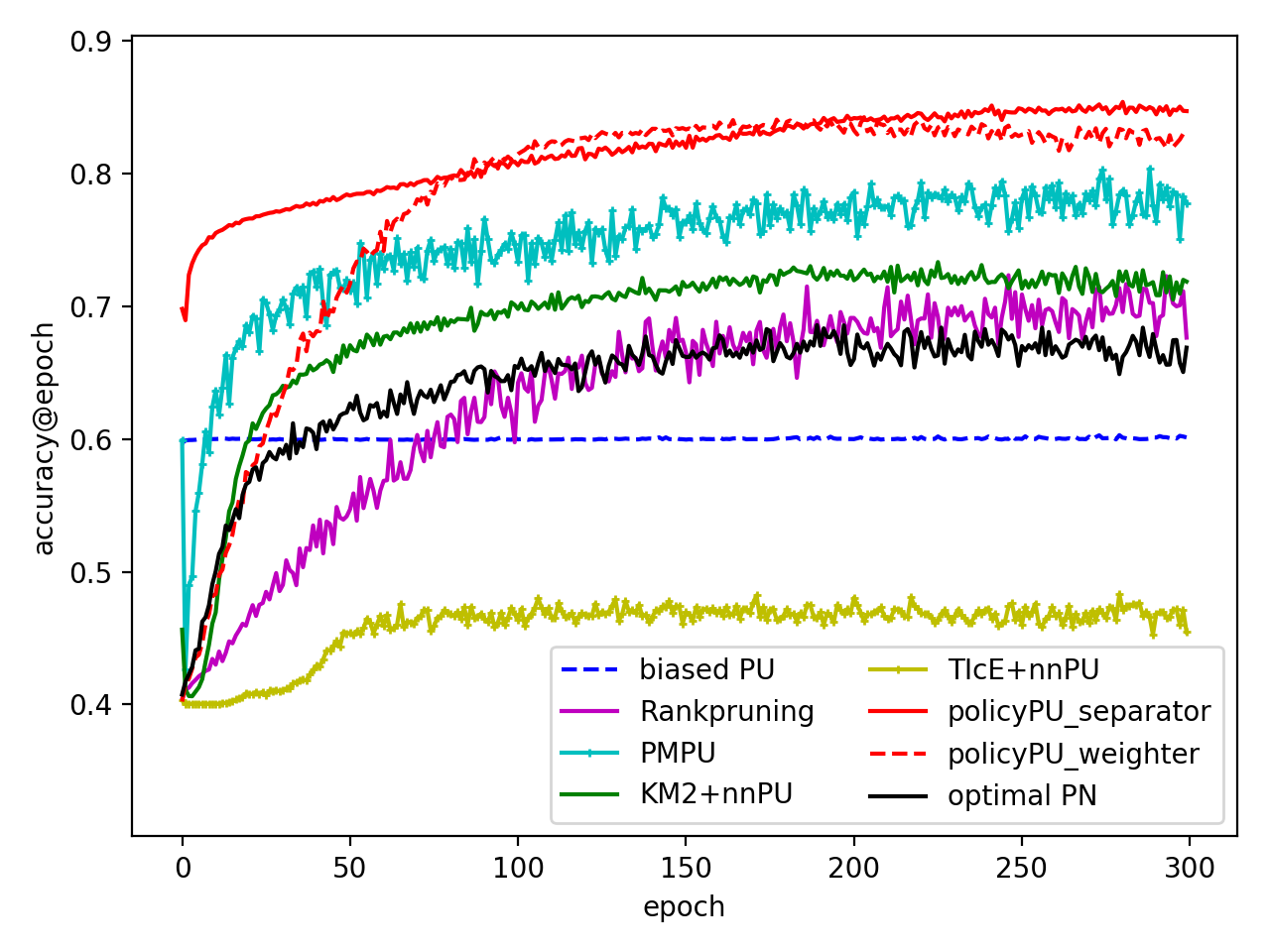}
  \end{subfigure}
  \caption{{\bf Accuracy comparison on {\em CIFAR-10} dataset.} \it The classifier is a 6-layer CNN model, and the policy is a 5-layer CNN model. The number of labeled examples is 300, 500 and 1,000 from the first row to the third row; The percentage of positive examples in the unlabeled data is 0.3, 0.5 and 0.7 from left to right.}
  \label{fig:cifar_cnn}
\end{figure*}


For the experiments on CIFAR-10, we train CNN models as classifiers and policies using the same architecture as the experiments on MNIST. Similarly, TIcE and KM2 algorithms are run first to make estimation before feeding the results to nnPU, separately. RankPruning eliminates label noises to create a relatively clean training dataset, and then builds a classification model on it.
We apply the same pre-training for our proposal and update policy once in 3 epochs to learn classifiers. The weight decay for classifiers is $2.0$, and for policy networks are $0.005$ and $1.0$, respectively. 

The experimental results are illustrated in Fig. \ref{fig:cifar_cnn}.
As shown, our framework is able to train classification models that generate higher accuracy compared to other algorithms. 
It is also recognized from the accuracy curve comparison that, our proposal sometimes even yields higher accuracy than the classifier trained on fully labeled PN data with the same parameter setting. 
We believe that if the true positive and negative examples in U dataset overlap near the decision boundary, the instance weights and even hard assignment given by the policy on these data may serve as an effective regularizer for the classifier.
It is an interesting phenomenon worth further investigation in the future.

We also observe interesting learning curves from the accuracy comparison on CIFAR-10, and in a few settings on MNIST as well. For some senarios, the policy network seems to be making inaccurate decisions for unlabeled examples at the beginning, yet quickly corrects itself after a few trials. It, in fact, reveals that the proposed interactive learning between the policy network and the classifier is effective for learning on PU datasets. Even if the policy network is inaccurate at the beginning of training, coherence rewards provided by the classifier would allow the policy network and targeted classifier to learn from each other and quickly rectify the policy.
The detailed comparison on ROC\_AUC, accuracy and PR\_AUC after 300-epoch training is presented in Table \ref{tab: cifar_300}, \ref{tab: cifar_500} and \ref{tab: cifar_1000}.

\begin{table}[t!]
\caption{{\bf Experiment results on {\em CIFAR-10} with CNN classifiers.} \it No. of labeled examples is 300. The percentage of P in U is 0.3, 0.5 and 0.7. ROC\_AUC, accuracy and PR\_AUC are shown from left to right for each percentage setting.
}
\begin{center}
\label{tab: cifar_300} 
\resizebox{\columnwidth}{!}{
\begin{tabular}{ l c  c c c |c c c| c c c }
\toprule

Model && \multicolumn{3}{c}{{\em 0.3}} & \multicolumn{3}{c}{{\em 0.5}} & \multicolumn{3}{c}{{\em 0.7}} \\
\cmidrule{1-2} \cmidrule{3-5} \cmidrule{6-8} \cmidrule{9-11} 

biased PU && 0.906 & 0.683 & 0.865  &  0.864 & 0.622 & 0.812 &  0.833 & 0.607 & 0.757 \\

TIcE+nnPU && 0.888 & 0.682 & 0.828  &  0.872 & 0.657 & 0.810 &   \bftab0.858 &0.646& \bftab0.784  \\

KM2+nnPU && 0.895 & 0.814  & 0.877  &  0.884 & 0.736 & 0.825  &   0.849 &0.691& 0.781 \\

RankPruning && 0.901 & 0.792 & 0.857   &   0.880  & 0.754 & 0.832   &   0.819 &0.577& 0.734 \\

PMPU && 0.874 & 0.774 & 0.808   &  0.849 & 0.721 & 0.767   &   0.833 &0.721& 0.743 \\
\cmidrule{1-2} \cmidrule{3-5} \cmidrule{6-8} \cmidrule{9-11}

policyPU\_separator && \bftab0.915 & \bftab0.835 & 0.877   &   \bftab0.886 & \bftab0.808 & \bftab0.842   &   0.847 &\bftab0.750& 0.780 \\

policyPU\_weighter && \bftab0.915 & 0.830 & \bftab0.880   &   0.875 & 0.798 & 0.831   &   0.818 &0.741& 0.742 \\
\cmidrule{1-2} \cmidrule{3-5} \cmidrule{6-8} \cmidrule{9-11}

optimal PN && 0.935 & 0.919 & 0.907  &  0.926 & 0.779 & 0.894   & 0.927 &0.623 &  0.895   \\
\bottomrule
\end{tabular}}
\end{center}
\end{table}

\begin{table}[thb]
\caption{{\bf Experiment results on {\em CIFAR-10} with CNN classifiers.} \it No. of labeled examples is 500. The percentage of P in U is 0.3, 0.5 and 0.7. ROC\_AUC, accuracy and PR\_AUC are shown from left to right for each percentage setting.
}
\begin{center}
\label{tab: cifar_500} 
\resizebox{\columnwidth}{!}{
\begin{tabular}{ l c  c c c |c c c| c c c }
\toprule
Model && \multicolumn{3}{c}{{\em 0.3}} & \multicolumn{3}{c}{{\em 0.5}} & \multicolumn{3}{c}{{\em 0.7}} \\
\cmidrule{1-2} \cmidrule{3-5} \cmidrule{6-8} \cmidrule{9-11} 

biased PU && 0.926 & 0.680 & 0.895  &  0.899 & 0.619 & 0.859   &  0.859 &0.607& 0.807 \\

TIcE+nnPU && 0.900 & 0.621 & 0.847  &  0.878 & 0.553 & 0.813   & 0.881 &0.510& 0.827  \\

KM2+nnPU && 0.915 & 0.803 & 0.871  &  0.903 & 0.720 & 0.851   &   \bftab0.885 &0.744& \bftab0.842 \\

RankPruning && 0.929 & 0.726 & 0.902   &   0.901 & 0.792 & 0.859   &   0.837 &0.584& 0.778 \\

PMPU && 0.880 & 0.749 & 0.812   &   0.880 & 0.748 & 0.806   &   0.865 &0.743& 0.788 \\
\cmidrule{1-2} \cmidrule{3-5} \cmidrule{6-8} \cmidrule{9-11}

policyPU\_separator && 0.930 & \bftab0.860 & 0.897   &   \bftab0.910 & \bftab0.834 & \bftab0.874   &   0.876 &\bftab0.787& 0.834 \\

policyPU\_weighter && \bftab0.935 & 0.859 & \bftab0.907    &   0.907 & 0.823 & 0.866   &   0.847 &0.770& 0.790 \\
\cmidrule{1-2} \cmidrule{3-5} \cmidrule{6-8} \cmidrule{9-11}

optimal PN && 0.949 & 0.874 & 0.926  &  0.945 & 0.800 & 0.921  & 0.939 &0.643 &  0.913   \\
\bottomrule
\end{tabular}}
\end{center}
\end{table}

\begin{table}[t!]
\caption{{\bf Experiment results on {\em CIFAR-10} with CNN classifiers.} \it No. of labeled examples is 1,000. The percentage of P in U is 0.3, 0.5 and 0.7. ROC\_AUC, accuracy and PR\_AUC are shown from left to right for each percentage setting.
}
\begin{center}
\label{tab: cifar_1000} 
\resizebox{\columnwidth}{!}{
\begin{tabular}{ l c  c c c |c c c| c c c }
\toprule
Model && \multicolumn{3}{c}{{\em 0.3}} & \multicolumn{3}{c}{{\em 0.5}} & \multicolumn{3}{c}{{\em 0.7}} \\
\cmidrule{1-2} \cmidrule{3-5} \cmidrule{6-8} \cmidrule{9-11} 

biased PU && 0.939 & 0.703 & 0.919  &  0.920 & 0.618 & 0.885   &  0.911 &0.602& 0.875 \\

TIcE+nnPU && 0.904 & 0.641 & 0.859  &  0.906 & 0.592 & 0.859   & 0.876 &0.443& 0.817  \\

KM2+nnPU && 0.924 & 0.811 & 0.893  &  0.917 & 0.745 & 0.875   &   \bftab0.918 &0.703& \bftab0.880 \\

RankPruning && 0.948 & 0.825 & 0.927   &   0.929 & 0.712 & \bftab0.898   &   0.905 &0.667& 0.861 \\

PMPU && 0.907 & 0.764 & 0.860   &   0.887 & 0.731 & 0.817   &   0.869 &0.762& 0.792 \\
\cmidrule{1-2} \cmidrule{3-5} \cmidrule{6-8} \cmidrule{9-11}

policyPU\_separator && 0.938 & 0.856 & 0.914   &   0.926 & 0.846 & 0.891  &   0.911 &\bftab0.833& 0.875 \\

policyPU\_weighter && \bftab0.951 & \bftab0.884 & \bftab0.933    &  \bftab 0.929 & \bftab0.853 & 0.897   &   0.902 &0.816& 0.858 \\
\cmidrule{1-2} \cmidrule{3-5} \cmidrule{6-8} \cmidrule{9-11}

optimal PN && 0.955 & 0.879 & 0.940  &  0.994 & 0.814 & 0.932  &  0.991 &0.650 &  0.919   \\
\bottomrule
\end{tabular}}
\end{center}
\end{table}

\begin{table}[ht!]
\caption{{\bf Experiment results on {\em UserTargeting} dataset.} \it The classifiers are two 6-layer MLPs ([d-100-50-50-30-1]), and they are paired with a 4-layer (left) and a 6-layer (right) policy networks.}
\begin{center}
\label{tab: prophet_mlp}
\resizebox{\columnwidth}{!}{
\begin{tabular}{ l c c c c c  c c c c}
\toprule
 & \multicolumn{4}{c}{{\em 4-layer policy}}& \multicolumn{4}{c}{{\em 6-layer policy}}\\
\cmidrule{3-5} \cmidrule{7-9} 
Model && ROC\_AUC & Accuracy & PR\_AUC && ROC\_AUC & Accuracy & PR\_AUC\\
\cmidrule{1-2} \cmidrule{3-5} \cmidrule{7-9} 
biased PU &&0.951 & 0.872 &  0.828 && 0.956 & 0.885 & 0.849\\ 
TIcE+nnPU &&0.957 & 0.895 &  0.841 && 0.958 & 0.895 & 0.849\\ 
KM2+nnPU &&0.957 & 0.896 & 0.846 && 0.955 & 0.896 & 0.847\\ 
RankPruning &&0.949 & 0.880 & 0.821 && 0.956 & 0.886 & 0.842\\ 
PMPU &&0.858 & 0.631 & 0.526 && 0.844 & 0.629 & 0.518\\ 
\cmidrule{1-2} \cmidrule{3-5} \cmidrule{7-9} 
policyPU\_separator &&\bftab 0.974 & \bftab 0.937 & \bftab 0.898 && \bftab0.972 & \bftab0.941 & \bftab 0.892\\
policyPU\_weighter && 0.971 & 0.914 & 0.885 && 0.969 & 0.919 & 0.883\\
\bottomrule
\end{tabular}}
\end{center}
\end{table}

\subsection{Experiments on the UserTargeting dataset}
We train two MLP classifiers with 1,000 labeled examples, and they are learned together with a 6-layer MLP and 4-layer MLP as policy networks, respectively.
A narrower architecture, [d-100-50-50-30-1], is used for the neural networks due to the dimension of user feature vector. The weight decay for our classifiers and policy networks are set to $2.0$ and $1e\mbox{-}4$, separately. Experiment results are summarized after running 1,000 epochs.

\begin{figure}[t!]
  \centering
  \hspace*{-1em}
  \begin{subfigure}[b]{0.48\columnwidth}
  	\includegraphics[width=\columnwidth]{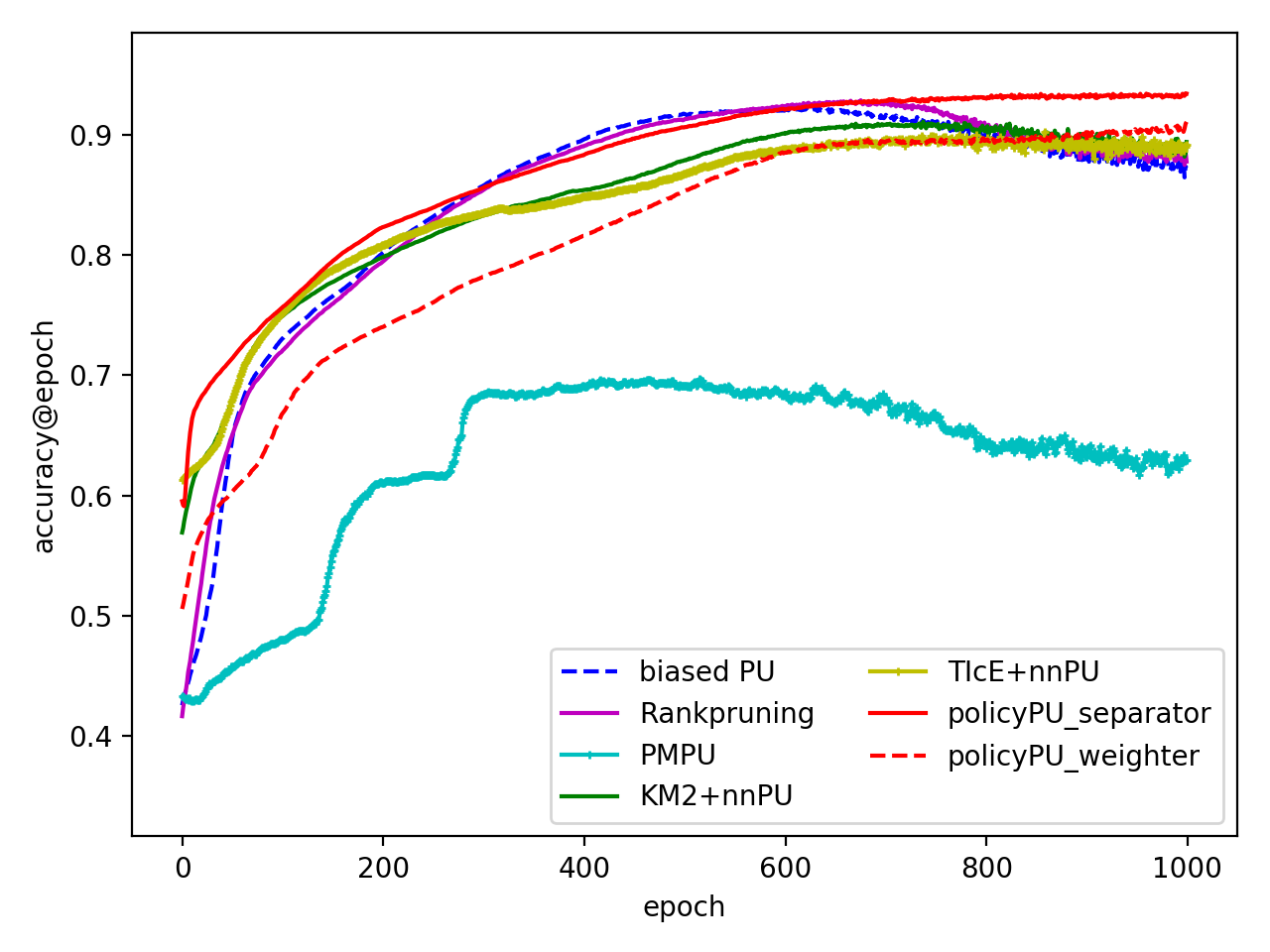}
  \end{subfigure}\hspace*{0em}
  \begin{subfigure}[b]{0.48\columnwidth}
  	\includegraphics[width=\columnwidth]{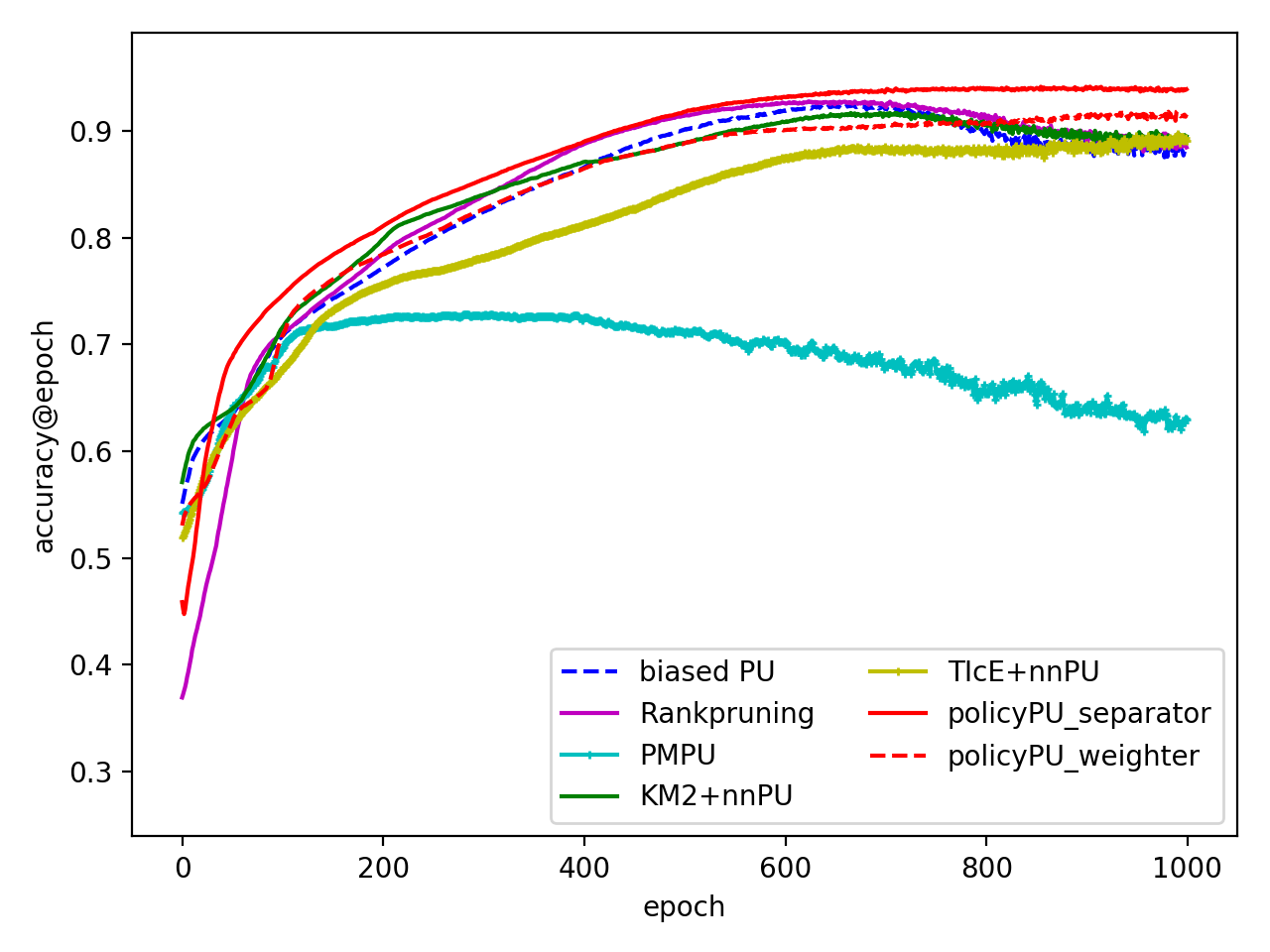}
  \end{subfigure}\hspace*{-1em}
  \caption{{\bf Accuracy comparison on {\em UserTargeting} dataset.} \it The classifier is 6-layer MLP, the paired policy network is a 4-layer MLP (left) and a 6-layer MLP (right). The number of labeled examples is 1,000.}
  \label{fig:prophet_result}
\end{figure}

The accuracy comparison is presented in Fig. \ref{fig:prophet_result}. Note that since UserTargeting dataset does not contain any true negative examples, there is no comparison to optimal PN learning in the experiment. 
For other baseline algorithms, we follow the same training procedure described for the experiments on MNIST and CIFAR-10.
As displayed, unfortunately PMPU struggles to have good performance this time, unlike on the other two datasets. We speculate that the reason is due to the fact that this user behavior dataset is noisier than MNIST and CIFAR-10. Hence, the calculation of the significant parameter, $\tau$, for PMPU may be severely impacted.

We recognize that it takes a bit longer for policy networks to learn consistent policy. We assume it is still because of the high noise level, which makes the policy learning converge slower. It is hard for policy to make quick decisions on how an unlabeled data instance should be used.
RankPruning is able to produce very promising results. Its pruning of likely mislabeled examples works well for this user behavior dataset.
Besides, both class prior estimation methods seem to have difficulty to accurately approximate true values. 
Meanwhile, biased PU learning turns out to be a strong baseline for this dataset as the assumption that the unlabeled instances being mostly negative may actually hold for this particular problem. Yet, eventually our classifiers can yield comparable performance. Another important observation is that our classifiers do not severely suffer from overfitting problem in the end.
The comparison on ROC\_AUC, accuracy and PR\_AUC are shown in Table \ref{tab: prophet_mlp}.

\subsection{Verification on policy learning}
\begin{figure}[t!]
  \centering
  \hspace*{-1em}
  \begin{subfigure}[b]{0.48\columnwidth}
  	\includegraphics[width=\columnwidth]{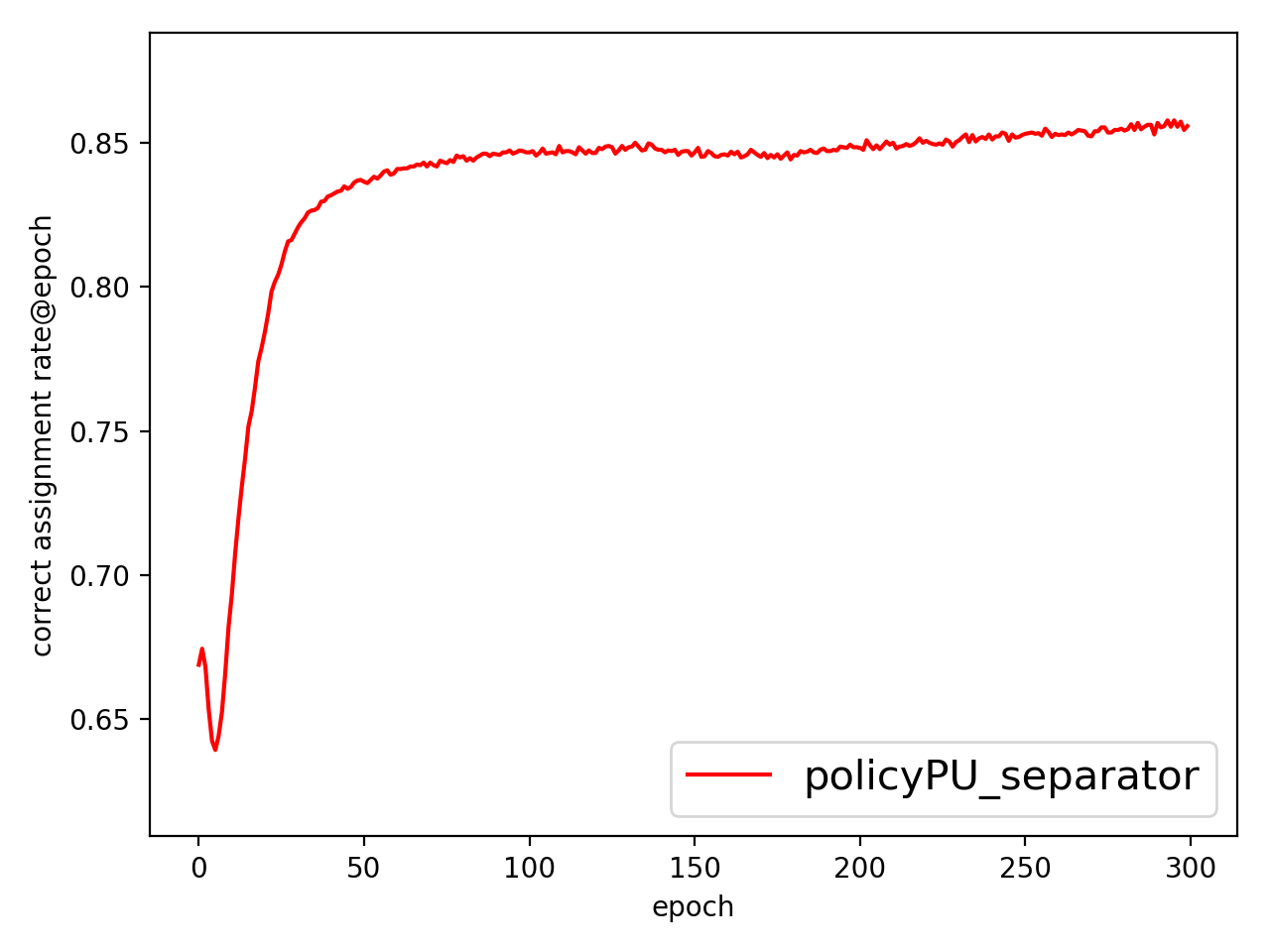}
  \end{subfigure}\hspace*{0em}
  \begin{subfigure}[b]{0.48\columnwidth}
  	\includegraphics[width=\columnwidth]{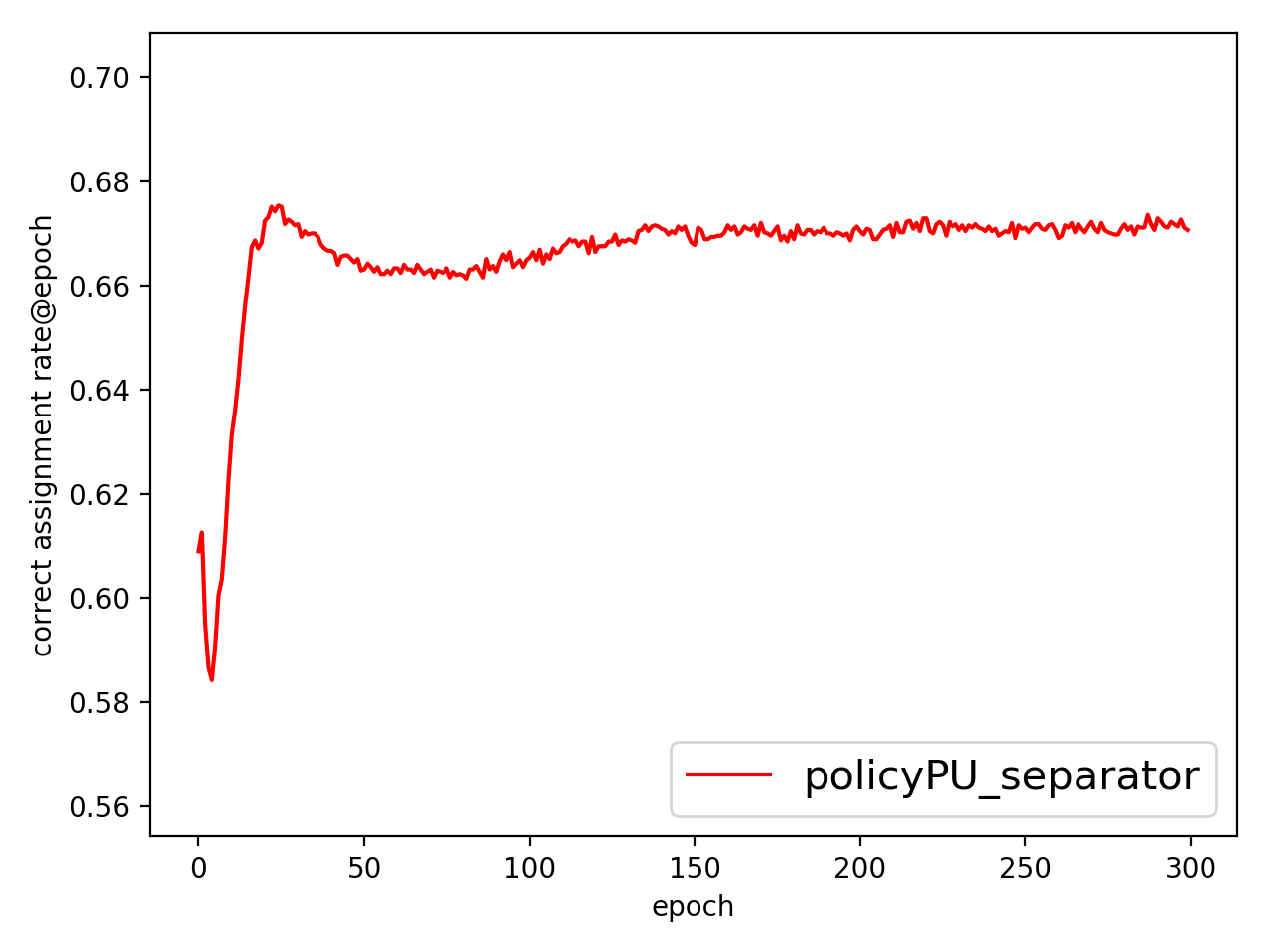}
  \end{subfigure}\hspace*{-1em}
  \caption{{\bf Correct assiginment rate of the unlabeled examples by the policy in \emph{Separator}.} \it The experiment is run with 300 labeled examples, and the percentage of positive in unlabeled data is 0.3. Experiment results on \emph{MNIST} (left) and on \emph{CIFAR-10} (right) are shown.}
  \label{fig:eta_validation}
\end{figure}

In this subsection, we demonstrate whether the policy is learning to improve its decision making on the unlabeled examples.
In our proposed interactive learning mechanism, the policy must update itself towards better policy during the training process in order to facilitate the classifier learning. 
As shown in Fig. \ref{fig:mnist_cnn} and Fig. \ref{fig:cifar_cnn}, the classification models learn to yield more accurate performance.
Here, we illustrate policy's decision making on the unlabeled data to verify if they are gradually getting better during the training as well.
Since the policy in \emph{Separator} directly makes a hard assignment which is more straightforward to understand, we use its results as examples for discussion. 
As indicated in Fig. \ref{fig:eta_validation}, the rate of correctly assigned data instances is getting better in the training process for both scenarios. 
The drop at the beginning, in fact, matches one of the observations elaborated in the experiment on CIFAR-10, that at first the policy is making wrong decisions. However, the interactive learning can correct it after a few epochs of trials.
We can see that the policy is indeed improving along with the classifier in overall. This observation is actually consistent with the theoretical analysis in \cite{ZZ18a}. 
They propose a generalized cross entropy loss and derive a bound of the optimal objective function value difference between using the dataset with true labels and using the dataset with noisy labels. The latter corresponds to the PU learning in our experiment. Further, it proves that as noise rate decreases, the optimal function value on a noisy dataset approaches to the one using a clean dataset.


\section{Conclusions}

This paper proposed a reinforcement learning framework, in which a policy network learns to update its assumptions of unlabeled examples, and a classifier that builds on the actions taken by the policy, makes predictions and generates rewards to guide the policy training. Compared to existing PU learning methods which rely on a pipeline to make estimations on unlabeled examples and to build a classifier, the interactive learning between the policy and the classifier in our proposed framework is able to make use of U data in a more effective manner, and train a more generalized classifier in an end-to-end fashion. Experimental results on three datasets demonstrate that the classifiers learned by our framework are able to yield performance improvement in terms of ROC\_AUC, accuracy and PR\_AUC.

\bibliography{main}
\bibliographystyle{ieeetr}
\end{document}